\documentclass[11pt]{article}

\usepackage[final]{acl}

\usepackage{times}
\usepackage{latexsym}
\usepackage[T1]{fontenc}
\usepackage[utf8]{inputenc}
\usepackage{inconsolata}

\usepackage{microtype}
\usepackage{url}
\usepackage{booktabs}
\usepackage{bm}
\usepackage{graphicx}
\usepackage{amsmath}
\usepackage{subcaption}
\usepackage{wrapfig}
\usepackage[table]{xcolor}
\usepackage{tikz}
\usetikzlibrary{fit}
\usepackage{amssymb}
\usepackage{array}
\usepackage{multirow}
\usepackage{makecell}
\usepackage{caption}
\usepackage{algorithm}
\usepackage{algpseudocode}
\usepackage[most]{tcolorbox}
\usepackage{nicematrix}
\usepackage{arydshln}

\definecolor{rice}{RGB}{250,245,232} 
\definecolor{ricegreen}{RGB}{232, 245, 218} 

\definecolor{darkblue}{rgb}{0, 0, 0.5}
\hypersetup{colorlinks=true, citecolor=darkblue, linkcolor=darkblue, urlcolor=darkblue}

\title{Listen to the Latents: Self-Correcting Speech Recognition in Large Audio Language Models Through Hidden-State Interactions}

\author{%
  Chan-Jan Hsu$^{\ddagger}$\quad Jaeyeon Kim$^{\ddagger}$\quad Chao-Han Huck Yang$^{\natural}$\thanks{Work done at NVIDIA}\quad \\
  \textbf{Shinji Watanabe}$^{\ddagger}$\textbf{\quad}
  \textbf{Hung-yi Lee}$^{\flat}$\textbf{\quad}
  \textbf{Carlos Busso}$^{\ddagger}$ \\
  $^{\ddagger}$Carnegie Mellon University\quad
  $^{\flat}$National Taiwan University\quad
  $^{\natural}$NVIDIA \\
  \texttt{chanjanh@andrew.cmu.edu, busso@cmu.edu} \\
}

\usepackage{tikz}
\usepackage{array}
\definecolor{NERGood}{HTML}{6BAED6}
\definecolor{NERBad}{HTML}{C6DBEF}
\definecolor{NonNERGood}{HTML}{FF7F0E}
\definecolor{NonNERBad}{HTML}{FDD0A2}

\begin{document}

\maketitle

\begin{abstract}
Recent automatic speech recognition (ASR) systems increasingly integrate large language models (LLMs) to leverage their semantic knowledge, either externally through logit fusion or internally through warm initialization. 
However, how to effectively combine these two strategies remains underexplored.
In this work, we refine warm-initialized LLM-based ASR models by leveraging their own pre-adaptation base LLMs, focusing on LoRA-adapted settings where the base LLM is preserved.
To achieve this, we propose Hybrid Search, a targeted correction strategy motivated by two observations.
First, interaction features that characterize the relationship between LLM-based ASR hidden states and base-LLM hidden states provide informative signals about a token's degree of semantic dependence. 
Second, selectively refining targeted tokens with high semantic dependence improves ASR performance far beyond naive global LLM-correction methods including rescoring and late fusion.
Our analysis suggests that, even after semantic knowledge transfer through warm initialization, LLM-based ASR models can still leverage their base LLM to further improve inference-time performance. Demo: \url{https://huggingface.co/spaces/Splend1dchan/Listen-To-The-Latent}

\end{abstract}

\section{Introduction}
Recent automatic speech recognition (ASR) models with billions of parameters trained on hundreds of thousands to millions of hours of audio have demonstrated strong performance across a wide range of scenarios \citep{whisper, canary, owsm, owsm31}. However, there remains a substantial data gap between ASR models and large language models (LLMs). For example, \texttt{Whisper-large-v3} was trained on approximately 5 million hours of speech, which roughly corresponds to tens of billions of text tokens in the paired transcriptions \cite{openai_whisper_large_v3}. This data scale is orders of magnitude smaller than the trillions of tokens used to train modern LLMs.
This discrepancy results in a semantic gap: LLMs possess significantly richer semantic representations and broader world knowledge than ASR models. Consequently, leveraging LLMs offers a promising direction for improving ASR performance~\citep{ma2023n}, particularly for words that depend more on semantic context, such as named entities and uncommon words.

\begin{figure*}[t]
  \centering
  \includegraphics[width=\textwidth]{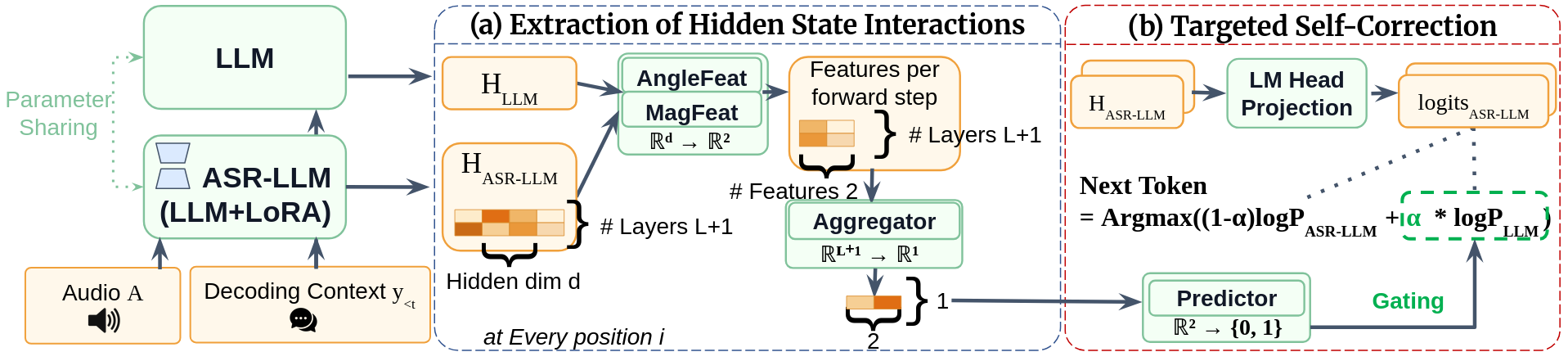}
\caption{ASR-LLM self-correction framework. (a) Section~\ref{sec:proposed}: Features are extracted from hidden-state interactions between the ASR-LLM and the LLM. (b) Section~\ref{sec:hid-correction}: These features govern a gating mechanism that decides whether to incorporate LLM log-probabilities into prediction.}
  \label{fig:pipeline}
\end{figure*}

Existing approaches typically coordinate LLMs with ASR systems in two major ways. One direction is to treat the entire system as a "single model" \citep{hsu2025let, mittal2024salsa, hori2025delayed}, where linguistic information from the LLM is typically incorporated through logit-level late fusion. Another line of work emphasizes the agentic role of LLMs in ASR \citep{yang2023generative,chen2023hyporadise,radhakrishnan2023whispering, wan2026speech}, where their instruction-following capabilities are used to refine hypotheses from the ASR model. However, these approaches fundamentally rely on correcting the outputs of a weaker ASR model rather than advancing the model's underlying recognition ability.

To leverage the benefits of LLMs in ASR, studies have explored directly integrating LLMs with the ASR backbone, which we refer to as ASR-LLM throughout this work. ASR-LLMs such as \texttt{Phi-4-Multimodal}, \texttt{SALM}, and \texttt{Qwen3-ASR} \citep{phi4_mm, salm, shi2026qwen3} are typically initialized from pretrained LLMs \citep{phi4_mm, shoeybi2019megatron, granite2024granite}. The pretrained LLM backbone is combined with a speech encoder and jointly trained on large-scale speech--language corpora. Models built in this way achieve performance competitive with, or superior to, ASR systems trained from scratch and post-hoc LLM correction methods \cite{srivastav2025openasrleaderboardreproducible}. This performance pattern suggests that LLM semantic knowledge transfers more effectively to ASR through warm initialization than through post-hoc correction.



While ASR-LLMs demonstrate strong empirical performance, it remains unclear how these models utilize the semantic and linguistic knowledge encoded within the integrated LLM. Prior work has leveraged output distributions or uncertainty signals to improve factuality during decoding \citep{chang2025real,hsu2025reducing,yang2026less}. 
Beyond output-level signals, layer-wise representation analysis has long shown that different layers encode complementary information \citep{peters2018deep}. 
More recent work interprets changes across layers as latent trajectories \citep{coe,chen2025reasoning, chuang2024dola}, which can provide useful signals for improving decoding.
For example, the aggregated angle and magnitude features between consecutive layers of latent trajectories has been shown to enable differentiation between correct and incorrect reasoning samples \cite{coe}. 
We extend prior studies by analyzing the geometric relationship between the hidden states of LoRA-adapted ASR-LLMs and those of their original base LLM. Our analysis shows that \textbf{cosine similarity} and \textbf{relative magnitude} between these hidden states provide informative signals for identifying tokens with high semantic dependence.

Building on this observation, we address a central question: can ASR-LLMs be further improved by reusing their original base LLMs for external guidance during inference?
We introduce \textbf{``Hybrid Search''}---a decoding algorithm that first uses hidden-state interactions to identify tokens whose features fall in regions associated with high semantic dependence, then applies targeted LLM correction to these tokens. This process is demonstrated in  Figure~\ref{fig:pipeline}.
Under our setting, Hybrid Search is fully \textbf{self-corrective} \cite{grill2020bootstrap,chou2025self}, requiring neither additional parameters nor further training. 
We focus our current evaluation on ASR, evaluating Hybrid Search under different computational budgets as both a beam-search-like decoding strategy \cite{hsu2025let} and a test-time scaling method \cite{snell2024scaling}. More broadly, the proposed framework may extend to other tasks or conditions where task-adapted models retain access to their original base LLM, such as speech translation and audio understanding \cite{hsu2025reducing, lin2026how}.
Experimental results on ASR show that Hybrid Search improves named-entity recognition over greedy search, beam search, and naive LLM-integration methods such as rescoring and late fusion, while maintaining a WER comparable to beam search.
These results suggest that, even after semantic knowledge transfer through warm initialization, an ASR-LLM can further leverage its original base LLM to improve inference-time performance.




\section{Background}

\textbf{LLM-based ASR.} Previous work has laid the foundation for warm-initializing LLM-based ASR across diverse architectural and training paradigms, including fully finetuned systems \cite{shi2026qwen3} and parameter-efficiently adapted models  \cite{phi4_mm,granite_speech,nvidia2025canaryqwen25b,salm}. These ASR-LLMs have been shown to be more effective than models trained from scratch \cite{canary,whisper,owsm}, as well as their LLM-corrected variants \cite{hsu2025let,chen2023hyporadise,lin-etal-2025-neko,wan2026speech}, on standard ASR benchmarks \cite{srivastav2025openasrleaderboardreproducible}. In this work, we investigate self-correction in LoRA-based ASR-LLMs through a representative case study of \texttt{Phi-4-Multimodal}, and further validate its generality with \texttt{Granite-4.0-1B-Speech}. Both models retain fully functional LLM backbones, enabling direct isolation of the LLM pathway.\footnote{For Phi-4-Multimodal, the backbone is Phi-4-Mini; for Granite-4.0-1B-Speech, the backbone is Granite-4.0-1B-Base.}
Furthermore, the ASR capability of both models has already been developed using a substantial amount of speech data (100k to 2.3 million hours), making further data scaling a less attractive path for addressing any shortcomings in semantic knowledge transfer.

\textbf{Named Entities in Speech Recognition.} Named entities and domain-specific rare words often carry disproportionate contextual importance in ASR, motivating extensive work on their recognition \cite{chen2023hyporadise}. 
Prior work has explored domain-specialized language models \cite{liu2021domain} and explicit domain tags \cite{liao2023zero} to improve recognition performance. Other approaches train embedded named-entity recognition modules within ASR systems \cite{ayache2025whisperner}.
Complementarily, SpeechIQ \cite{wan2025speechiq} addresses the limited emphasis that word error rate (WER) places on semantically consequential recognition errors by computing LLM-based similarity scores from embedding representations \cite{liu2024evaluating, jiang2024scaling}.
Following this line of work, we report named-entity error rate (NE-ER) in addition to WER to better characterize the semantic gains introduced by LLM-based correction.



\section{Hidden-State Interactions Reveal Token Semantic Dependence}
\label{sec:proposed}


\subsection{Extraction of Hidden State Interactions}
\label{sec:extract-hid}

\textbf{Formalization.} An ASR-LLM model with $L$ layers can be represented as a composition of ordered submodules:
\begin{equation}
f = f_{\mathrm{head}} \circ f_L \circ \cdots f_l \cdots \circ f_1 \circ f_{\mathrm{emb}} \, 
\label{eq:f}
\end{equation}
where the embedding module $f_{\mathrm{emb}}$ converts input tokens into $d$-dimensional embeddings. The intermediate layers $\{f_l\}_{l=1}^{L}$ successively transform these representations, and the classification head $f_{\mathrm{head}}$ maps the final hidden state into the vocabulary space $V$ to produce the output prediction.

We define the hidden states as the collection of layer-wise representations before the classification head. For simplicity, we denote the embedding layer $f_{\mathrm{emb}}$ as $f_0$. Given an input token sequence $\mathbf{I}$, we define the hidden-state vector $\mathbf{h}_l^i$ at token position $i$ as the output of the partial composition $f_l \circ \cdots \circ f_1 \circ f_0(\mathbf{I})^i$. We then aggregate these vectors across layers to obtain the layer-wise hidden states for position $i$:

\begin{equation}
\mathbf{H}^i = \big[(\mathbf{h}_0^i)^\top; \cdots; (\mathbf{h}_L^i)^\top\big] \in \mathbb{R}^{(L+1)\times d}
\end{equation}

In our setting, the model operates in a speech--text regime, where hidden states vary with the input modality. We denote by $\mathbf{H}_{\mathrm{ASR\text{-}LLM}}$ the hidden states produced by the ASR-LLM during ASR decoding at a given step, and by $\mathbf{H}_{\mathrm{LLM}}$ the hidden states obtained by running the base LLM on the corresponding text-only input.

\begin{equation}
\mathbf{H}_{\mathrm{ASR\text{-}LLM}} = f(\mathbf{I}_{<i}; A; \{\phi_{\mathrm{LLM}}, \phi_{\mathrm{LoRA}}\})
\end{equation}
\begin{equation}
\mathbf{H}_{\mathrm{LLM}} = f(\mathbf{I}_{<i}; \{\phi_{\mathrm{LLM}}\})
\end{equation}

Here, $\mathbf{H}_{\mathrm{ASR\text{-}LLM}}$ is conditioned on both the audio encoder output $A$ and the already decoded context plus the prompt  $\mathbf{I}_{<i}$, under the active ASR-LLM parameters $\{\phi_{\mathrm{LLM}}, \phi_{\mathrm{LoRA}}\}$. In contrast, $\mathbf{H}_{\mathrm{LLM}}$ is obtained by running the underlying LLM on only text context $\mathbf{I}_{<i}$ using only $\phi_{\mathrm{LLM}}$.
To analyze hidden-state interactions, we pair each speech-conditioned trajectory $\mathbf{H}_{\mathrm{ASR\text{-}LLM}}$ with its text-only counterpart $\mathbf{H}_{\mathrm{LLM}}$, computed under the same text context $\mathbf{I}_{<i}$.
\begin{figure*}[t]
\centering
\small

\begin{subfigure}[t]{0.9\textwidth}
\centering
\includegraphics[width=\linewidth]{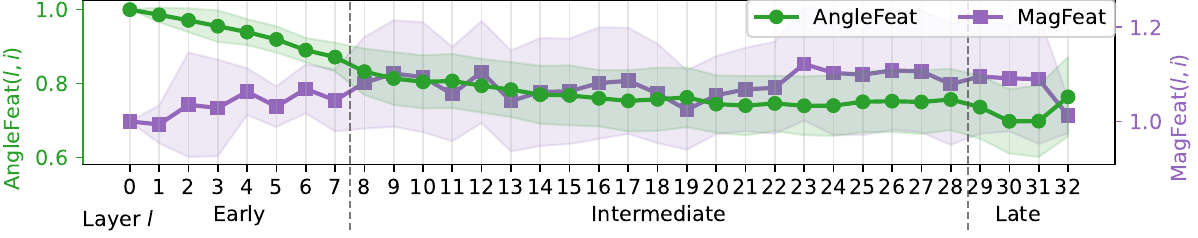}
\caption{Layer-wise averages and standard deviations of
$\mathrm{AngleFeat}$ and $\mathrm{MagFeat}$.}
\label{fig:gaussian_layerwise}
\end{subfigure}

\vspace{6pt}

\begin{subfigure}[t]{\textwidth}
\centering
\newcommand{\panelw}{0.27\textwidth}

\begin{tabular}{@{}>{\centering\arraybackslash}p{\textwidth}@{}}

\textbf{Intermediate Features:}
$f_{\mathrm{inter}}(i)
=
\frac{1}{3}
\sum_{l \in \operatorname{Top3}_{10 \le l \le 28}(f)}
f(l,i),\quad
f \in \{\mathrm{AngleFeat},\mathrm{MagFeat}\}$
\\[4pt]

\toprule

\begin{tabular}{@{}
>{\centering\arraybackslash}m{\panelw}
||
>{\centering\arraybackslash}m{\panelw}
>{\centering\arraybackslash}m{\panelw}
@{}}
\multirow{2}{*}{\textbf{Single-Token Words}} &
\multicolumn{2}{c@{}}{\textbf{Multi-Token Words}} \\
&
\textbf{Starting Token} &
\textbf{Continuation Token(s)}
\end{tabular}
\\[4pt]

\begin{tabular}{@{}
>{\centering\arraybackslash}m{\panelw}
||
>{\centering\arraybackslash}m{\panelw}
>{\centering\arraybackslash}m{\panelw}
@{}}
\includegraphics[width=\panelw]
{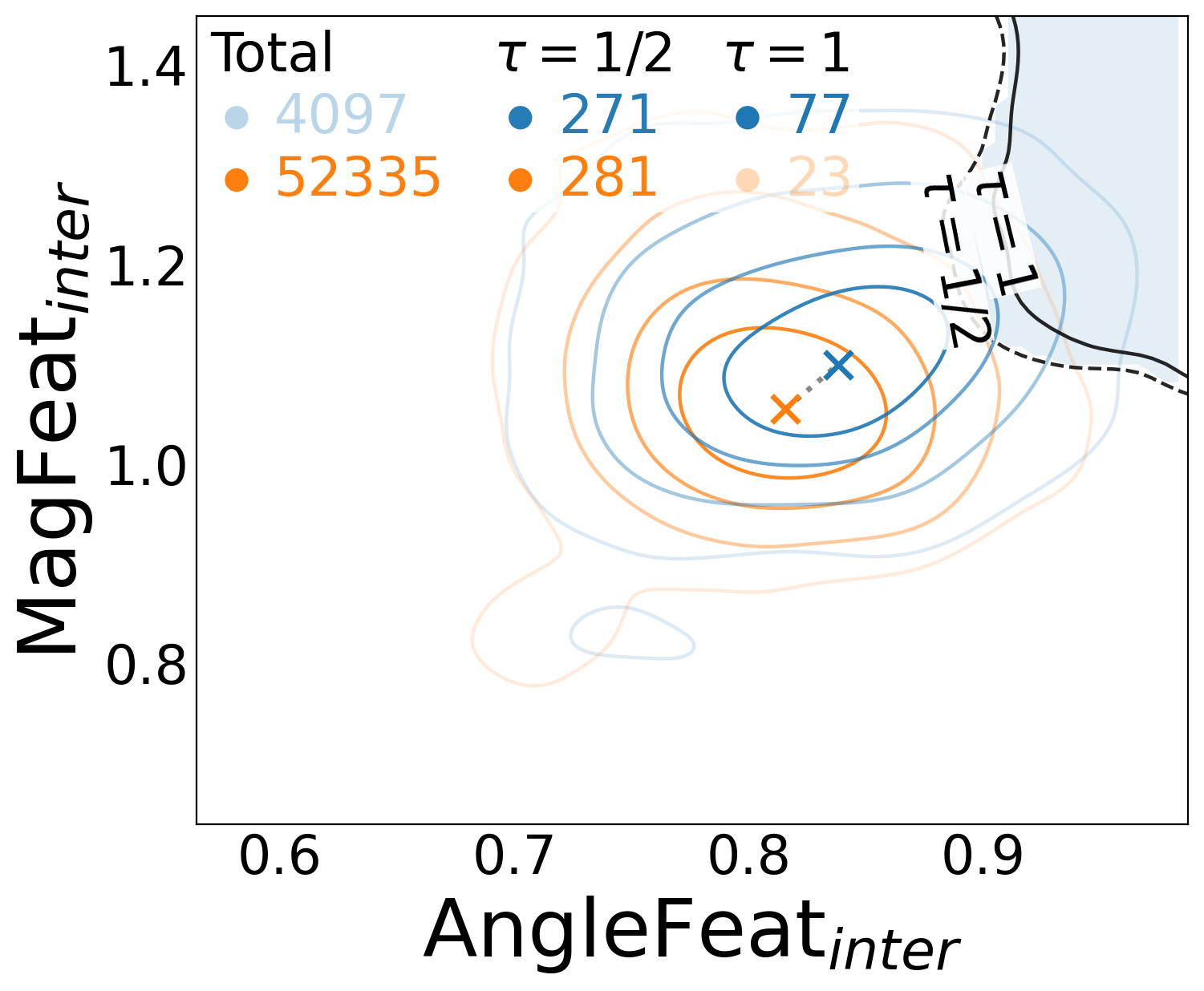}
&
\includegraphics[width=\panelw]
{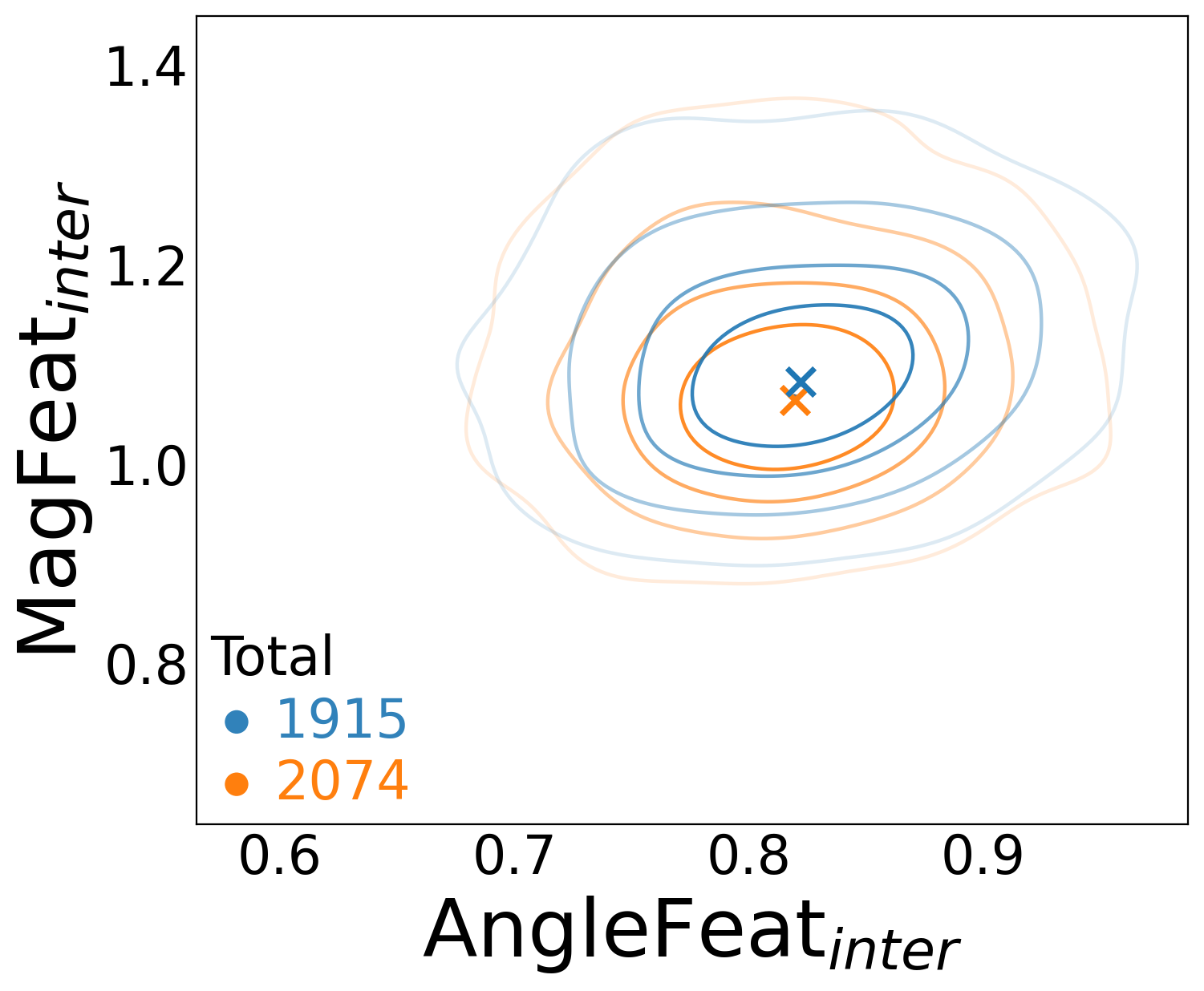}
&
\includegraphics[width=\panelw]
{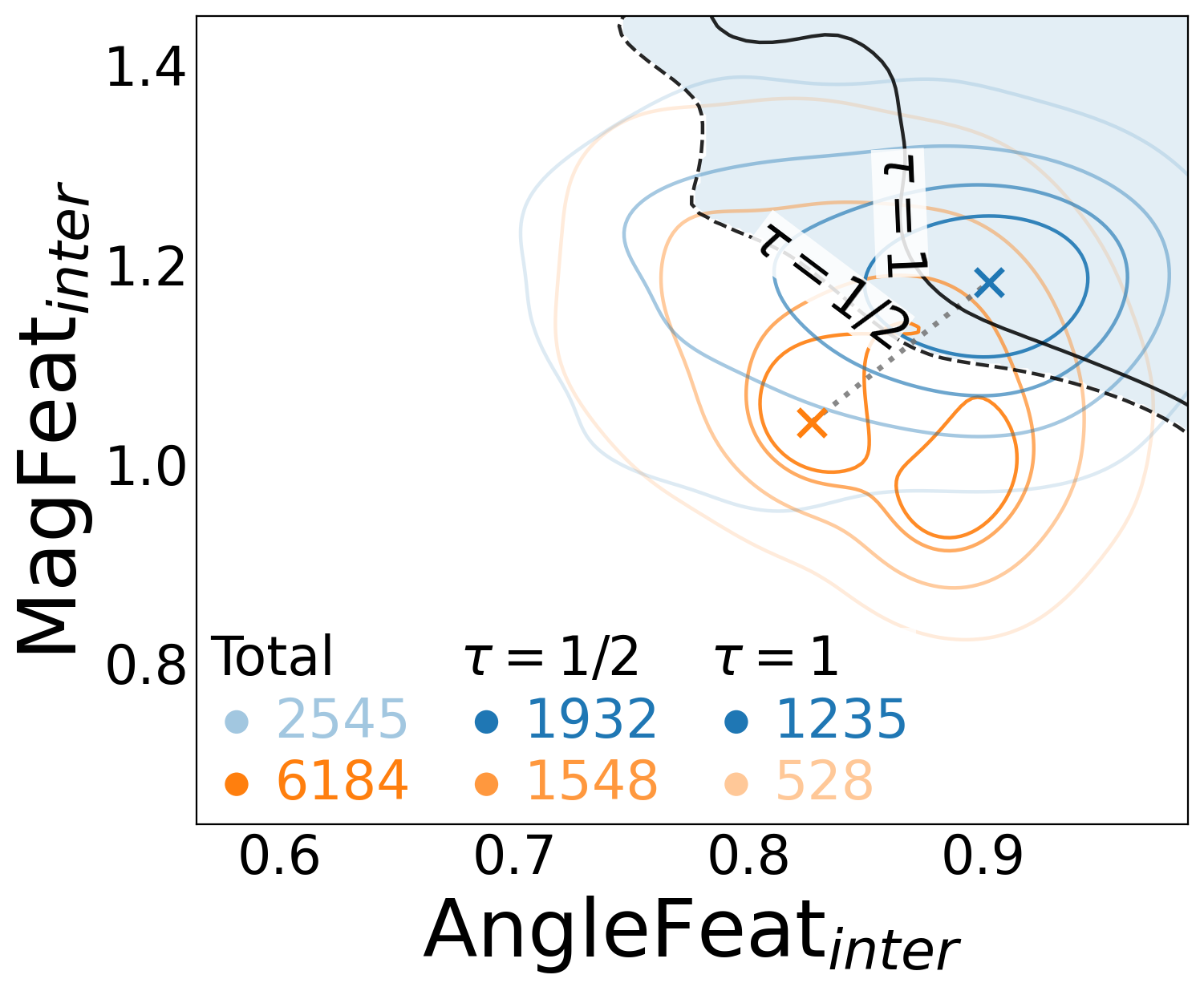}
\end{tabular}
\\[6pt]

\textbf{Late-Layer Features:}
$f_{\mathrm{late}}(i)=f(31,i),\quad
f \in \{\mathrm{AngleFeat},\mathrm{MagFeat}\}$
\\[4pt]

\toprule

\begin{tabular}{@{}
>{\centering\arraybackslash}m{\panelw}
||
>{\centering\arraybackslash}m{\panelw}
>{\centering\arraybackslash}m{\panelw}
@{}}
\multirow{2}{*}{\textbf{Single-Token Words}} &
\multicolumn{2}{c@{}}{\textbf{Multi-Token Words}} \\
&
\textbf{Starting Token} &
\textbf{Continuation Token(s)}
\end{tabular}
\\[4pt]

\begin{tabular}{@{}
>{\centering\arraybackslash}m{\panelw}
||
>{\centering\arraybackslash}m{\panelw}
>{\centering\arraybackslash}m{\panelw}
@{}}
\includegraphics[width=\panelw]
{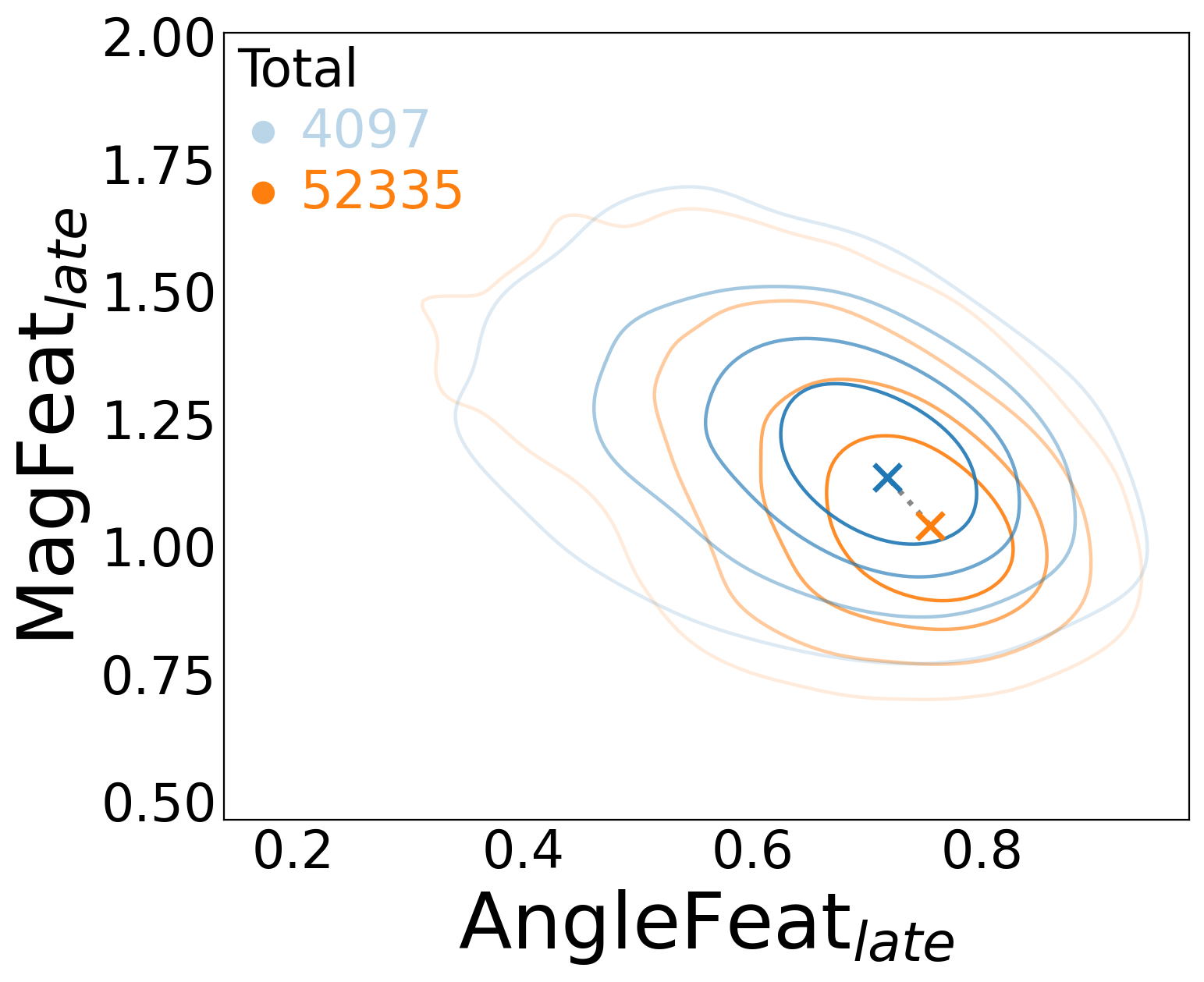}
&
\includegraphics[width=\panelw]
{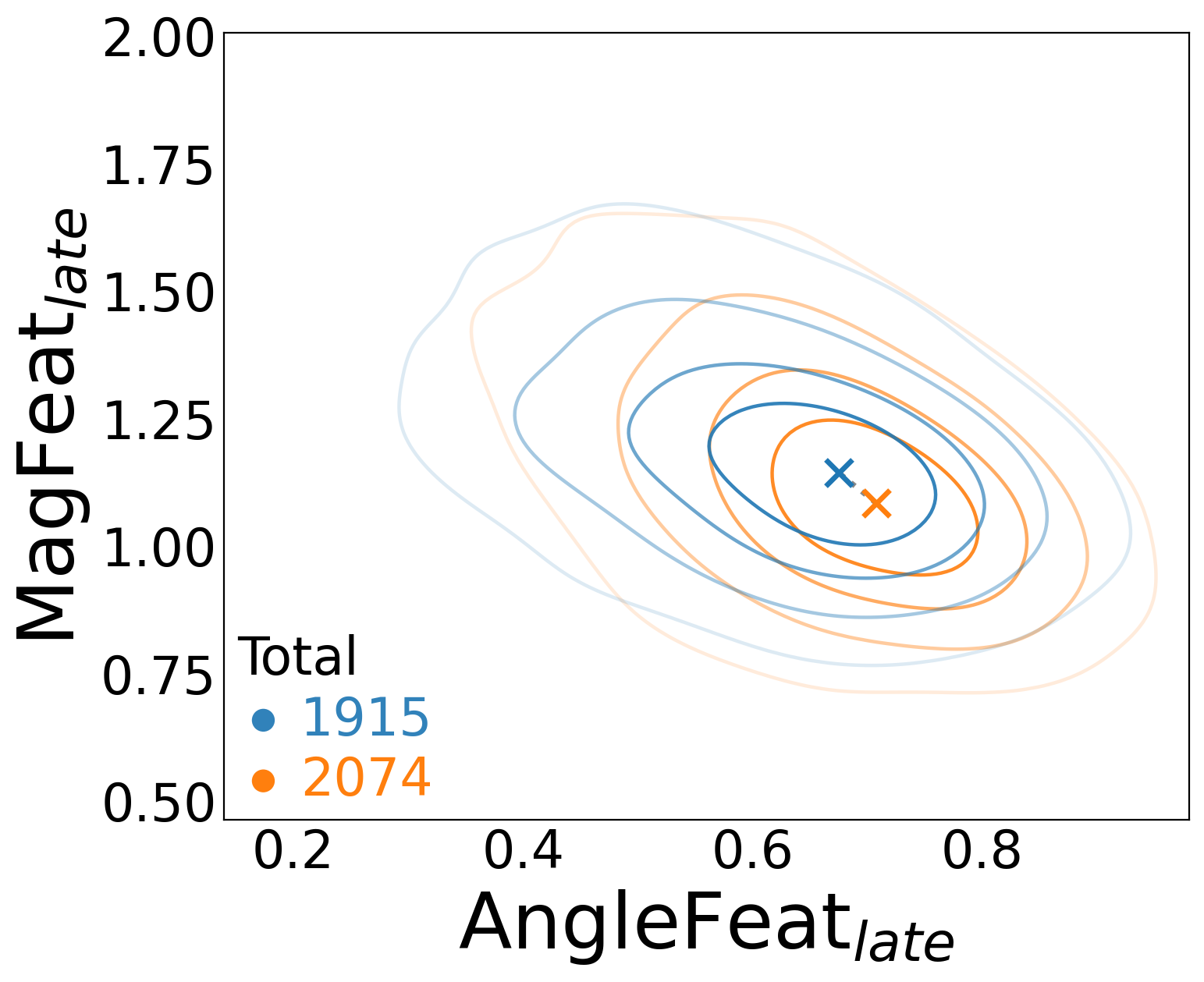}
&
\includegraphics[width=\panelw]
{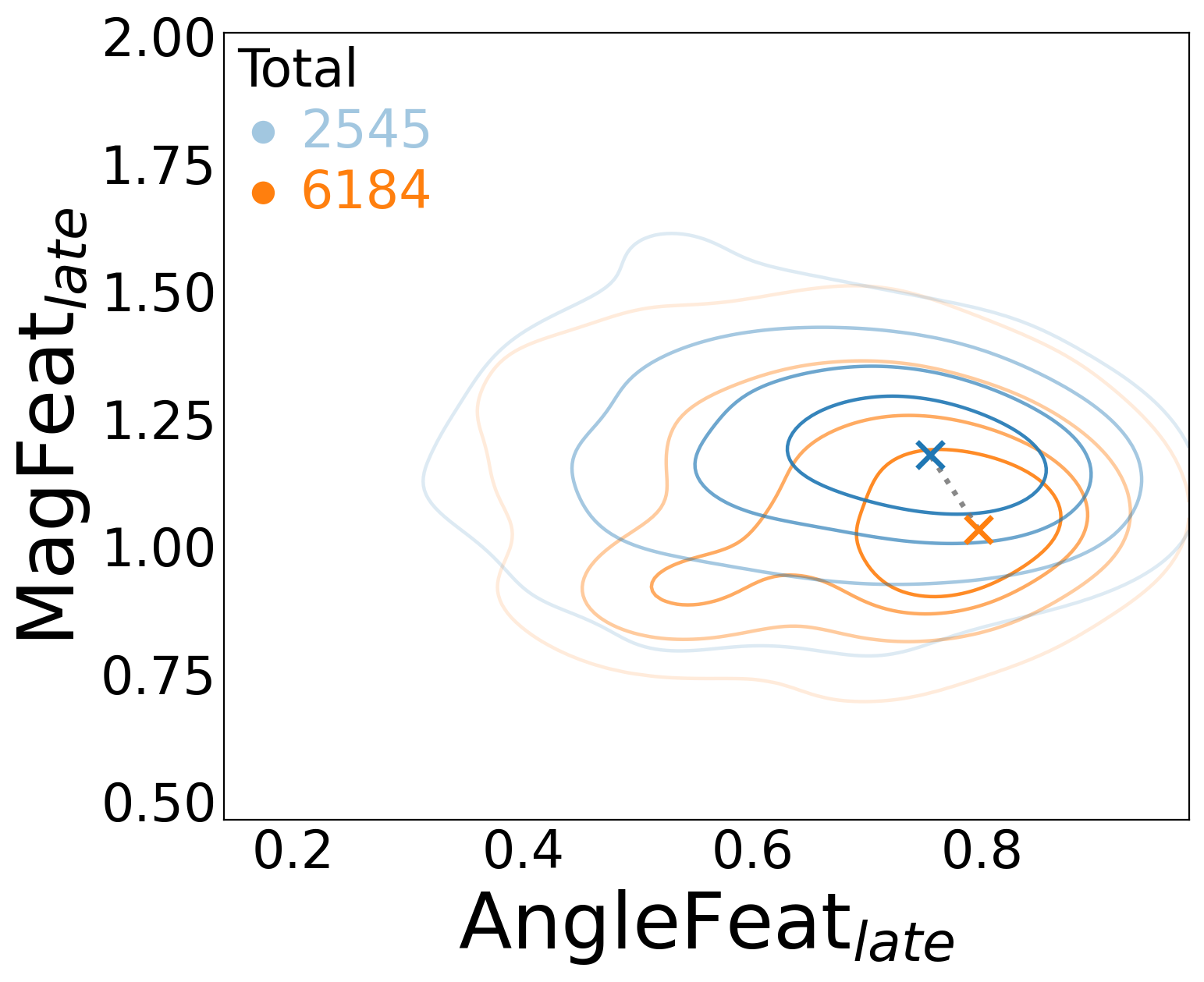}
\end{tabular}
\\

\addlinespace[4pt]

\textcolor[HTML]{FF7F0E}{\rule[0.6ex]{1.8em}{3pt}}
~\textcolor{black}{Non-named-entity tokens}
\qquad\qquad
\textcolor[HTML]{1F77B4}{\rule[0.6ex]{1.8em}{3pt}}
~\textcolor{black}{Named-entity tokens}
\\

\bottomrule

\end{tabular}

\caption{
Gaussian distributions of hidden-state interaction features for
\textbf{\textcolor[HTML]{1F77B4}{named-entity}} and
\textbf{\textcolor[HTML]{FF7F0E}{non-named-entity}} tokens.
The upper row aggregates intermediate-layer features, while the lower
row directly selects the features at layer 31.
}
\label{fig:gaussian_distribution}
\end{subfigure}

\caption{
\textit{Top:} Average values and standard deviations of
$\mathrm{AngleFeat}(l,i)$ and $\mathrm{MagFeat}(l,i)$ across tokens
reveal distinct layer-wise phases.
\textit{Bottom:} Hidden-state interaction feature distributions for
\textbf{\textcolor[HTML]{1F77B4}{named-entity}} and
\textbf{\textcolor[HTML]{FF7F0E}{non-named-entity}} tokens.
Intermediate-layer aggregation reveals a clear distribution shift,
while the late-layer features at layer 31 exhibit a further shift
toward lower cosine similarity and higher relative norm for
named-entity tokens.
}
\label{fig:gaussian}

\end{figure*}


\textbf{Feature Extraction.}
Given paired hidden states from the ASR-LLM and the base LLM
($\mathbf{H}_{\mathrm{ASR\text{-}LLM}}$ and $\mathbf{H}_{\mathrm{LLM}}$),
the representation associated with each prediction token remains high-dimensional,
with size $2 \times (L+1) \times d$.
To obtain compact and interpretable descriptors, prior work reduces the hidden dimension $d$ by deriving \textbf{angle} and \textbf{magnitude} features between inter-layer representation trajectories~\citep{coe}.
Unlike prior work, we compute them by comparing ``cross-modal'' hidden states
$\mathbf{H}_{\mathrm{LLM}}$ and $\mathbf{H}_{\mathrm{ASR\text{-}LLM}}$
at corresponding layers.
This design follows from our goal of characterizing how the pre-adapted $\mathbf{H}_{\mathrm{LLM}}$ are transformed into the post-adapted $\mathbf{H}_{\mathrm{ASR\text{-}LLM}}$.
Thus, each layer serves as a matched reference point for measuring ASR-induced representational change.
For each layer $l$ and position $i$, we define these features as follows:
\begin{equation}
\mathrm{AngleFeat}(l,i)
:= \cos(\mathbf{h}^{l,i}_{\mathrm{ASR\text{-}LLM}},
       \mathbf{h}^{l,i}_{\mathrm{LLM}})
\label{eq:angfeat}
\end{equation}
\begin{equation}
\mathrm{MagFeat}(l,i)
= \|\mathbf{h}^{l,i}_{\mathrm{ASR\text{-}LLM}}\|_2
/ \|\mathbf{h}^{l,i}_{\mathrm{LLM}}\|_2.
\label{eq:magfeat}
\end{equation}

Under this formulation, each token is characterized by 2 scalar descriptors at each layer. In the following subsection, we further aggregate these layer-wise features over selected layers to obtain exactly 2 scalar features for each predicted token in the speech recognition task. This compact parameterization allows us to fit Gaussian distributions over different token categories, such as whether a token is semantically grounded, and to quantify distributional shifts across these categories.  



\subsection{How Hidden State Interactions Reveal Token Semantic Dependence}
\label{sec:hid-sem}
\textbf{Analysis Setup.} We aim to examine how $\textrm{AngleFeat}$ and $\textrm{MagFeat}$ characterize semantic reliance. To operationalize ``semantic dependence,'' we treat named-entity tokens as a representative subset, motivated by empirical evidence from prior work \cite{chen2023hyporadise}. Conversely, we use non-named-entity tokens as a proxy for tokens whose prediction is more strongly grounded in acoustic evidence. We perform this analysis on the \texttt{YODAS} dataset \citep{li2023yodas}, a CC-licensed audio corpus with substantial acoustic and linguistic diversity.
Named-entity tokens are identified using a BERT-based NER tagger\footnote{\href{https://huggingface.co/dslim/bert-base-NER-uncased}{dslim/bert-base-NER-uncased}}.
We select \texttt{Phi-4-Multimodal} as the ASR-LLM model, whose base LLM is \texttt{Phi-4-Mini} \citep{phi4_mm}. 
We apply teacher forcing using the reference transcription as input, yielding $2 \times N \times (L+1)$ features for each sample, where $N$ denotes the sequence length. These per-sample features are best visualized by heatmaps shown in  Appendix~\ref{apx:heat_map}.

We require another feature size reduction $(L+1)\rightarrow1$ for distributional analysis. To obtain proper aggregations on the layer dimension, we look into the macro trajectories of $\mathrm{AngleFeat}$ and $\mathrm{MagFeat}$ displayed at the top of Figure~\ref{fig:gaussian}.
We categorize the layers into early (0-7), intermediate (8-28), and late (29-32) phases based on changes in behavior observed in the feature trajectories. 
We structure the following discussion around these 3 layer phases to examine potential aggregation strategies.

\textbf{Early Layers, Layers 0--7:} Figure~\ref{fig:gaussian}(a) shows that both $\mathrm{AngleFeat}$ and $\mathrm{MagFeat}$ remain close to 1 in the early layers, indicating that the ASR-LLM hidden states deviate only minimally from the corresponding LLM hidden states. The token-level standard deviations indicated by the shaded regions are also small, suggesting that there is little room for different token categories to exhibit distinguishable trajectory patterns.

\textbf{Intermediate Layers, Layers 8--28:} The middle portion of Figure~\ref{fig:gaussian}(a) shows larger standard deviations across tokens, suggesting that feature values vary substantially depending on token identity and context. 
Manual inspection shows that named-entity tokens have higher $\mathrm{AngleFeat}$ and $\mathrm{MagFeat}$ within some of the intermediate layers, forming peak-like patterns. We thus derive ``peak'' features from intermediate layers by taking the top-3 mean of $\mathrm{AngleFeat}$ and $\mathrm{MagFeat}$ from layers 10 to 28. 
We refer to the aggregated features as $\mathrm{AngleFeat}_{inter}$ and $\mathrm{MagFeat}_{inter}$.
We then project each token onto a 2D feature plane using its peak feature values, and estimate the Gaussian feature distributions of named-entity and non-named-entity tokens, as visualized in Figure~\ref{fig:gaussian}(b).
From Figure~\ref{fig:gaussian}(b), we observe a distributional shift between the Gaussian centroids of named-entity and non-named-entity tokens for single-token words and continuation tokens. This suggests that the two token types follow distinct latent trajectories in the ASR-LLM.
In particular, named-entity tokens occupy the region characterized by high $\mathrm{AngleFeat}$ and high $\mathrm{MagFeat}$. 
We observe that the largest distributional discrepancy is primarily driven by continuation tokens in multi-token words, which are often collocative with the preceding token; for example, predicting ``-tras'' after ``orches'' to form the word ``orchestras.''
The ASR-LLM leverages the $\mathbf{H}_\mathrm{LLM}$ trajectory by aligning $\mathbf{H}_\mathrm{ASR\text{-}LLM}$ more closely with it in angular space during the intermediate layers. We demonstrate the generality of this behavior in Appendix~\ref{apx:granite} with \texttt{Granite-4.0-1B-Speech}.





\textbf{Late Layers, Layers 29--32:} In Figure~\ref{fig:gaussian}(a), late layers are characterized by a further decrease in $\mathrm{AngleFeat}$. When isolating layer 31, we find that named-entity tokens shift toward lower $\mathrm{AngleFeat}$ and higher $\mathrm{MagFeat}$ (Figure~\ref{fig:gaussian}(b)). Since this effect is weaker than the shift observed in the intermediate layers, we do not use it in our method. We include further analysis in Appendix~\ref{apx:late_feat}.


\textbf{Summary.} Our analysis shows that \textbf{multi-token named entities} exhibit decoding trajectories that are clearly distinct from those of non-named entities. This finding echoes the effect of tokenization bias \citep{lesci2025causal}, where probabilities for multi-token words can be miscalibrated relative to those for single-token words. In the following section, we examine how deviations in latent trajectories affect ASR correctness and how these insights can be used to improve recognition performance.
\begin{figure*}[t]
\centering
\scriptsize
\renewcommand{\arraystretch}{1.6}
\setlength{\tabcolsep}{5pt}
\arrayrulecolor{gray!45}

\scalebox{1}{%
\begin{tabular}{|l|
c|c|c|
>{\columncolor{rice}}c|
>{\columncolor{rice}}c|
>{\columncolor{rice}}c|
>{\columncolor{rice}}c|
>{\columncolor{rice}}c|
c|c|c|c|}
\hline
$\mathrm{AngleFeat}_{inter}$
& 0.83 & 0.80 & 0.77
& \multicolumn{1}{c|}{\tikz[remember picture,baseline=(n1.base)]\node (n1) {0.90};}
& \multicolumn{1}{c|}{\tikz[remember picture,baseline=(n2.base)]\node (n2) {\textbf{0.95}};}
& \multicolumn{1}{c|}{\tikz[remember picture,baseline=(n3.base)]\node (n3) {\textbf{0.90}};}
& \multicolumn{1}{c|}{0.86}
& \multicolumn{1}{c|}{}
& 0.78 & 0.81 & 0.83 & 0.85 \\
\hline

$\mathrm{MagFeat}_{inter}$
& 1.07 & 1.19 & 1.10
& \multicolumn{1}{c|}{\tikz[remember picture,baseline=(n4.base)]\node (n4) {1.10};}
& \multicolumn{1}{c|}{\tikz[remember picture,baseline=(n5.base)]\node (n5) {\textbf{1.18}};}
& \multicolumn{1}{c|}{\tikz[remember picture,baseline=(n6.base)]\node (n6) {\textbf{1.13}};}
& \multicolumn{1}{c|}{1.04}
& \multicolumn{1}{c|}{}
& 1.07 & 1.11 & 1.06 & 1.05 \\
\hline


& \multicolumn{3}{c|}{\textbf{Non Targeted}} &
\multicolumn{5}{c|}{\cellcolor{rice}\textbf{Targeted}} &
\multicolumn{4}{c|}{\textbf{Non Targeted}} \\
\hline

\textbf{Tokens}
& \texttt{\_Here} & \texttt{\_was} & \texttt{\_Bela}
& \texttt{\_Kar} & \texttt{ol} & \texttt{yi} & \texttt{,} & \texttt{(\_not)}
& \texttt{\_not} & \texttt{\_just} & \texttt{\_upset} & \texttt{\_that} \\[-2pt]

& & &
&
& \texttt{oly} & \texttt{,} & \texttt{(\_not)} & \texttt{$\varnothing$}
& & & & \\[-2pt]

& & &
& \texttt{\_Car}
& \texttt{oli} & \texttt{,} & \texttt{(\_not)} & \texttt{$\varnothing$}
& & & & \\



\hline
\end{tabular}%
}

\begin{tikzpicture}[remember picture, overlay]
\node[
    draw=red,
    rounded corners=3pt,
    very thick,
    inner xsep=4pt,
    inner ysep=3pt,
    fit=(n2)(n3)(n5)(n6)
] {};

\end{tikzpicture}

\arrayrulecolor{black}
\caption{Visualization of our proposed Hybrid Search. When features lie in the selected regions (red border), beam search is initiated (shaded region), backtracking to the beginning-of-word token. Upon reaching a consensus non-targeted beginning-of-word token (\texttt{"\_not"}), beams are rescored and decoding reverts to greedy search. The symbol $\varnothing$ denotes subsequent generated tokens that are omitted.}
\label{fig:hybrid}
\end{figure*}

\section{Hidden State Interactions Enable ASR Self-Correction}
\label{sec:hid-correction}
\subsection{Targeted Self-Correction}
\label{sec:target-correction}

\textbf{Formulation.} Prior late-fusion error-correction methods interpolate ASR and LLM probabilities for next-token selection \citep{hsu2025let,uadf}. When applied to the ASR-LLM setting, this gives:
\begin{equation}
(1-\alpha) \log p_{\textrm{ASR\text{-}LLM}} + \alpha \log p_{\textrm{LLM}}
\label{eq:logp}
\end{equation}
where $\log p_{\textrm{M}}$ denotes the log-probabilities of the next token produced by the model $\textrm{M}$, and $\alpha$ controls the interpolation weight.
While LLM contributions can correct token predictions that rely heavily on semantic context, the lack of audio input also makes LLM-based fusion susceptible to hallucinations when the decoded token is acoustically dominant \citep{hsu2025let}. Therefore, we explore a targeted correction framework that applies LLM probability fusion only to targeted, semantically grounded tokens. The log-probability score is correspondingly updated to a piecewise function:
\begin{equation}
\begin{cases}
(1-\alpha)\log p_{\textrm{ASR\text{-}LLM}} + \alpha\log p_{\textrm{LLM}}, 
& \text{if targeted},\\[2pt]
\log p_{\textrm{ASR\text{-}LLM}}
& \text{otherwise.}
\end{cases}
\label{eq:mixture}
\end{equation}

\textbf{Targeted Token Selection.} We identify named-entity-dense regions from the Gaussian distributions introduced in Section~\ref{sec:hid-sem}. 
These regions define a targeted-token selector $\mathcal{T}(\cdot)$:
\begin{equation}
\mathcal{T}(x_i)=
\mathbf{1}\left[
\frac{P_{\mathrm{NE}}(\mathbf{z}_i)}
     {P_{\mathrm{nonNE}}(\mathbf{z}_i)}
\ge \tau
\right],
\label{eq:target-selector}
\end{equation}
where $\mathbf{z}_i=(\operatorname{AngleFeat}_i,\operatorname{MagFeat}_i)$ denotes the two-dimensional feature representation of token $x_i$, and $P_{\mathrm{NE}}$ and $P_{\mathrm{nonNE}}$ are the Gaussian density estimates from the YODAS analysis for named-entity and non-named-entity tokens, respectively. 
The threshold $\tau$ controls the permissiveness of the targeted region.

We visualize two Gaussian density-ratio contours for each scenario in Figure~\ref{fig:gaussian}: the equi-density boundary ($\tau=1$), where the estimated named-entity and non-named-entity densities are equal, and a more permissive 1:2 density-ratio boundary ($\tau=\frac{1}{2}$), which expands the selected region to include more named-entity tokens at the cost of lower precision. For multi-token words, we only apply the boundary to continuation tokens, as starting tokens do not exhibit clear separation between named-entity and non-named-entity tokens. The decision boundaries identify an upper-right region where named-entity tokens are densely concentrated.  For interpretability, we perform word-level selection by expanding each targeted token to the full word span containing it. This means that a multi-token word is selected if any of its continuation tokens is targeted.
We note that the NER model or ground-truth entity information is no longer needed after calibration, and the inference algorithm depends only on the hidden-state interaction features $\mathbf{z}_i$.




\subsection{Self-Correction Algorithm} 
\label{sec:self-correct}

\begin{table}[t]
\centering
\small
\setlength{\tabcolsep}{5pt}
\renewcommand{\arraystretch}{1.0}

\begin{tabular}{l|cc}
\toprule
\textbf{Experiment}
& \begin{tabular}[c]{@{}c@{}}\textbf{Decoding}\\\textbf{Passes}\end{tabular}
& \begin{tabular}[c]{@{}c@{}}\textbf{Relative}\\\textbf{Cost}\end{tabular} \\
\midrule

Greedy Search
& One-Pass
& $1\times$ \\

\rowcolor{ricegreen}
\quad + Hybrid Search
& 
& \\[-1pt]

\rowcolor{ricegreen}
\quad\phantom{+ }(Decoding)
& \multirow{-2}{*}{One-Pass}
& \multirow{-2}{*}{$1.4\times^{*}$} \\

Beam Search
& One-Pass
& $5\times$ \\

\rowcolor{rice}
\quad + Hybrid Search
&
& \\[-1pt]

\rowcolor{rice}
\quad\phantom{+ }(Error Correction)
& \multirow{-2}{*}{Two-Pass}
& \multirow{-2}{*}{--} \\

\bottomrule
\multicolumn{3}{@{}l@{}}{%
    \parbox{0.8\columnwidth}{%
        \small
        $^{*}$Amortized effective beam size, averaged over tokens
        decoded with beam search and greedy search.
        (See Section~\ref{sec:Complexity})%
    }%
} \\

\end{tabular}

\caption{Hybrid Search for Decoding provides an efficient alternative within the family of beam-search methods, whereas Hybrid Search for Error Correction requires a second pass but further improves upon Beam Search (Table~\ref{tab:verbose_hybrid}).}
\label{tab:hybrid_efficiency}
\end{table}

We now instantiate Equation~\ref{eq:mixture} as a decoding algorithm. A challenging aspect of this formula lies in the instability of cross-beam comparisons between targeted tokens and non-targeted tokens, as their scores are on different scales. We propose \textbf{Hybrid Search}, a decoding algorithm that restricts comparisons to hypotheses with aligned targeted and non-targeted token groups. Hybrid Search is a mixture between greedy search and beam search, and  conducts beam search (beam size $B$) only for targeted words.

The full decoding process switches state between different beam sizes according to the token type. The transition from beam size 1 to $B$ upon encountering a targeted token is trivial. To reduce the beam size $B$ to 1, we continue decoding until all beams converge on the same non-targeted token. We then roll back any beams that may have decoded past  the converged token and select the best hypothesis from the resulting candidates.
Hybrid Search greatly reduces decoding cost relative to beam search, since most decoding steps remain in the greedy phase. We note that this process is made possible by targeting only a subset of tokens for expansion. Figure~\ref{fig:hybrid} illustrates hybrid search in action through explored decoding traces. We provide the full procedure in Algorithm~\ref{alg:hybrid-search} in the Appendix~\ref{apx:hybrid_algo}.


\section{Experimental Settings}

We evaluate Hybrid Search in two settings. First, we look into \textbf{Hybrid Search for Decoding}, directly implementing the procedure in Section~\ref{sec:self-correct}. By applying beam expansion only at selected positions, the method incurs a cost between greedy search and full beam search while remaining a one-pass decoding algorithm. Second, we explore \textbf{Hybrid Search for Error Correction}, adopting a two-pass procedure that starts from the best beam-search hypothesis and applies Hybrid Search as a refinement step. For non-targeted positions, we use teacher forcing on the beam-search hypothesis rather than rerunning decoding. The relationship between the two distinct Hybrid Search methods and their decoding complexity is shown in Table~\ref{tab:hybrid_efficiency}.

We evaluate Hybrid Search on \texttt{Phi-4-Multimodal} using the Open ASR Leaderboard benchmark \citep{srivastav2025openasrleaderboardreproducible}, following its standard setup of seven audio subsets drawn from diverse sources: AMI \citep{ami}, Earnings22 \citep{earnings22}, GigaSpeech \citep{gigaspeech}, LibriSpeech \citep{librispeech}, SPGISpeech \citep{spgispeech}, TED-LIUM \citep{tedlium}, and VoxPopuli \citep{wang2021voxpopuli}. In addition, we include Common Voice \citep{ardila2020common} as an additional evaluation set to assess robustness across diverse speakers. We report both WER and named-entity error rate (NE-ER). NE-ER is calculated as the number of unmatched named-entity words divided by the total number of named-entity words, where named entities are identified using the BERT-based NER tagger in Section~\ref{sec:hid-sem}.
Following beam search conventions, we set the beam size of Hybrid Search to $B=5$, and use an interpolation weight of $\alpha=0.2$. 
We report results with $\tau = \frac{1}{2}$ by default, and additionally evaluate $\tau = 1$ for Hybrid Search for Error Correction.
\begin{table*}[t]
\centering
\small
\renewcommand{\arraystretch}{1}
\caption{WER and NE-ER across the Open ASR Leaderboard and Common Voice (in \%). \textbf{Top:} Phi-4-Multimodal Beam Search outperforms the best prior LLM-corrective methods that incorporate beam search internally on most benchmarks.
\textbf{Middle and Bottom:} Hybrid Search applied to Greedy Search provides a cost-effective alternative to Beam Search, recovering 25\% of its WER improvement and 96\% of its NE-ER improvement. When applied to Beam Search, Hybrid Search further reduces average NE-ER by 3.3\% while maintaining comparable WER. In contrast, naive refinement methods such as rescoring and late fusion noticeably degrade WER. Values below the refinement results report two-sided exact McNemar test $p$-values against the Beam Search baseline; significant results ($p<0.05$) are shown in teal and non-significant results in gray.
}
\resizebox{1\textwidth}{!}{
\begin{tabular}{l*{8}{c}|c}
\toprule
&
\multicolumn{7}{c}{Open ASR Leaderboard} &
\multirow{2}{*}{\shortstack[c]{Common\\Voice}} &
\multirow{2}{*}{Avg.} \\
\cmidrule(lr){2-8}
& AMI & Earn. & Giga. & Libri. & SPGI. & TED. & Vox. &  &  \\
\midrule

\multicolumn{10}{c}{\textbf{Word Error Rate}} \\
\midrule

\textbf{(A)} GER $\Rightarrow$ Parakeet \cite{chen2023hyporadise}
& 22.91 & -- & 12.10 & 4.84 & 3.98 & 6.09 & 7.49 & -- & -- \\

\textbf{(B)} Neko Qwen1.5-MoE \cite{lin-etal-2025-neko}
& 12.60 & 11.82 & 9.95 & 2.32 & 1.94 & 3.20 & 5.80 & -- & -- \\

\textbf{(C)} Speech-Hands $\rightleftharpoons$ Parakeet \cite{wan2026speech}
& 11.20 & -- & 11.10 & 3.18 & 2.16 & 4.37 & 6.02 & -- & -- \\

\addlinespace[2pt]
\cmidrule(lr){1-10}

\textit{Best of LLM-corrective (A--C)} (Beam Search)
& 11.20 & 11.82 & 9.95 & \textbf{2.32} & \textbf{1.94} & 3.20 & 5.80 & -- & -- \\

\addlinespace[2pt]

Phi-4-Multimodal (Beam Search)
& \textbf{11.11} & \textbf{9.39} & \textbf{9.07} & 3.64 & 2.41 & \textbf{2.64} & \textbf{5.74} & 6.91 & 6.36 \\

\midrule
\multicolumn{10}{c}{
\textbf{Word Error Rate}
\quad {\small (McNemar $p$-value below)}
} \\
\midrule

Phi-4-Multimodal (Greedy Search)
& 11.38 & 9.68 & 9.19 & 3.82 & 2.46 & 2.72 & 5.97 & 7.33 & 6.57 \\

\rowcolor{ricegreen}
+ Hybrid Search ($\tau=\frac{1}{2}$)
& 11.38 & 9.73 & 9.07 & 3.70 & 2.48 & 2.68 & 5.87 & 7.20 & 6.51 \\

\rowcolor{ricegreen}
\quad - LLM Correction
& 11.35 & 9.73 & 9.09 & 3.72 & 2.47 & 2.70 & 5.83 & 7.09 & 6.50 \\

\addlinespace[2pt]
\cdashline{1-10}

Phi-4-Multimodal (Beam Search)
& 11.11 & \textbf{9.39} & 9.07 & 3.64 & 2.41 & 2.64 & 5.74 & 6.91 & 6.36 \\

\textcolor{gray}{+ Rescoring}
&
\textcolor{gray}{12.50} &
\textcolor{gray}{10.08} &
\textcolor{gray}{9.22} &
\textcolor{gray}{3.67} &
\textcolor{gray}{2.48} &
\textcolor{gray}{2.71} &
\textcolor{gray}{6.11} &
\textcolor{gray}{6.89} &
\textcolor{gray}{6.71} \\

\textcolor{gray}{Late Fusion}
&
\textcolor{gray}{12.79} &
\textcolor{gray}{12.05} &
\textcolor{gray}{12.07} &
\textcolor{gray}{5.89} &
\textcolor{gray}{5.05} &
\textcolor{gray}{5.03} &
\textcolor{gray}{8.32} &
\textcolor{gray}{11.68} &
\textcolor{gray}{9.11} \\

\addlinespace[2pt]
\cdashline{1-10}

Phi-4-Multimodal (Beam Search)
& 11.11 & \textbf{9.39} & 9.07 & 3.64 & 2.41 & 2.64 & 5.74 & 6.91 & 6.36 \\

\rowcolor{rice}
\raisebox{1.2ex}{+ Hybrid Search ($\tau=1$)}
&
\shortstack[c]{11.11\\{\scriptsize\textcolor{gray}{1.00}}}
&
\shortstack[c]{9.40\\{\scriptsize\textcolor{gray}{1.00}}}
&
\shortstack[c]{\textbf{9.05}\\{\scriptsize\textcolor{teal}{$<1e\text{-}5$}}}
&
\shortstack[c]{3.64\\{\scriptsize\textcolor{gray}{.749}}}
&
\shortstack[c]{2.41\\{\scriptsize\textcolor{teal}{$<1e\text{-}3$}}}
&
\shortstack[c]{2.62\\{\scriptsize\textcolor{gray}{.125}}}
&
\shortstack[c]{5.73\\{\scriptsize\textcolor{gray}{.688}}}
&
\shortstack[c]{6.89\\{\scriptsize\textcolor{gray}{.051}}}
&
\shortstack[c]{6.36\\{\scriptsize\textcolor{teal}{$<1e\text{-}8$}}}
\\[3pt]

\rowcolor{rice}
\raisebox{1.2ex}{+ Hybrid Search ($\tau=\frac{1}{2}$)}
&
\shortstack[c]{11.12\\{\scriptsize\textcolor{gray}{.451}}}
&
\shortstack[c]{9.40\\{\scriptsize\textcolor{gray}{.620}}}
&
\shortstack[c]{\textbf{9.05}\\{\scriptsize\textcolor{teal}{.007}}}
&
\shortstack[c]{3.63\\{\scriptsize\textcolor{gray}{.671}}}
&
\shortstack[c]{\textbf{2.40}\\{\scriptsize\textcolor{teal}{.016}}}
&
\shortstack[c]{\textbf{2.59}\\{\scriptsize\textcolor{teal}{.004}}}
&
\shortstack[c]{5.74\\{\scriptsize\textcolor{gray}{.864}}}
&
\shortstack[c]{\textbf{6.88}\\{\scriptsize\textcolor{gray}{.071}}}
&
\shortstack[c]{\textbf{6.35}\\{\scriptsize\textcolor{teal}{$<1e\text{-}4$}}}
\\[3pt]

+ Hybrid Search ($\tau=0$)
& 11.16 & 9.56 & 9.06 & 3.64 & 2.41 & 2.65 & 5.83 & 6.94 & 6.41 \\

\midrule
\multicolumn{10}{c}{
\textbf{Named-Entity Error Rate}
\quad {\small (McNemar $p$-value below)}
} \\
\midrule

Phi-4-Multimodal (Greedy Search)
& 26.57 & 31.78 & 21.64 & 23.27 & 15.46 & 9.91 & 5.15 & 22.59 & 19.55 \\

\rowcolor{ricegreen}
+ Hybrid Search ($\tau=\frac{1}{2}$)
& 24.71 & 29.44 & 20.31 & 22.21 & 15.20 & 9.03 & 4.91 & 21.23 & 18.38 \\

\rowcolor{ricegreen}
\quad - LLM Correction
& 25.33 & 30.20 & 20.77 & 22.53 & 15.36 & 9.03 & 4.91 & 21.66 & 18.72 \\

\cdashline{1-10}
\addlinespace[2pt]

Phi-4-Multimodal (Beam Search)
& 24.71 & 30.20 & 20.82 & \textbf{22.12} & 14.48 & 8.32 & 4.86 & 20.83 & 18.29 \\

\textcolor{gray}{+ Rescoring}
&
\textcolor{gray}{25.47} &
\textcolor{gray}{29.49} &
\textcolor{gray}{20.11} &
\textcolor{gray}{22.90} &
\textcolor{gray}{12.78} &
\textcolor{gray}{7.61} &
\textcolor{gray}{4.63} &
\textcolor{gray}{20.10} &
\textcolor{gray}{17.89} \\

\textcolor{gray}{Late Fusion}
&
\textcolor{gray}{24.79} &
\textcolor{gray}{30.27} &
\textcolor{gray}{21.14} &
\textcolor{gray}{22.72} &
\textcolor{gray}{14.77} &
\textcolor{gray}{8.32} &
\textcolor{gray}{5.34} &
\textcolor{gray}{22.49} &
\textcolor{gray}{18.73} \\

\cdashline{1-10}
\addlinespace[2pt]

Phi-4-Multimodal (Beam Search)
& 24.71 & 30.20 & 20.82 & \textbf{22.12} & 14.48 & 8.32 & 4.86 & 20.83 & 18.29 \\

\rowcolor{rice}
\raisebox{1.2ex}{+ Hybrid Search ($\tau=1$)}
&
\shortstack[c]{24.71\\{\scriptsize\textcolor{gray}{1.00}}}
&
\shortstack[c]{29.89\\{\scriptsize\textcolor{gray}{1.00}}}
&
\shortstack[c]{20.46\\{\scriptsize\textcolor{teal}{$<1e\text{-}6$}}}
&
\shortstack[c]{22.21\\{\scriptsize\textcolor{gray}{1.00}}}
&
\shortstack[c]{14.15\\{\scriptsize\textcolor{teal}{$<1e\text{-}7$}}}
&
\shortstack[c]{7.79\\{\scriptsize\textcolor{gray}{.25}}}
&
\shortstack[c]{4.72\\{\scriptsize\textcolor{gray}{.25}}}
&
\shortstack[c]{20.48\\{\scriptsize\textcolor{teal}{$<1e\text{-}4$}}}
&
\shortstack[c]{18.05\\{\scriptsize\textcolor{teal}{$<1e\text{-}16$}}}
\\[3pt]

\rowcolor{rice}
\raisebox{1.2ex}{+ Hybrid Search ($\tau=\frac{1}{2}$)}
&
\shortstack[c]{24.32\\{\scriptsize\textcolor{gray}{.125}}}
&
\shortstack[c]{29.82\\{\scriptsize\textcolor{gray}{.383}}}
&
\shortstack[c]{\textbf{20.11}\\{\scriptsize\textcolor{teal}{$<1e\text{-}11$}}}
&
\shortstack[c]{22.53\\{\scriptsize\textcolor{gray}{.298}}}
&
\shortstack[c]{\textbf{13.09}\\{\scriptsize\textcolor{teal}{$<1e\text{-}19$}}}
&
\shortstack[c]{7.26\\{\scriptsize\textcolor{gray}{.070}}}
&
\shortstack[c]{\textbf{4.43}\\{\scriptsize\textcolor{teal}{.012}}}
&
\shortstack[c]{\textbf{19.96}\\{\scriptsize\textcolor{teal}{$<1e\text{-}10$}}}
&
\shortstack[c]{\textbf{17.69}\\{\scriptsize\textcolor{teal}{$<1e\text{-}37$}}}
\\[3pt]

+ Hybrid Search ($\tau=0$)
& 23.78 & 29.74 & 20.33 & 22.49 & 13.68 & 8.14 & 4.67 & 20.43 & 17.91 \\

\bottomrule
\end{tabular}
}
\label{tab:verbose_hybrid}
\end{table*}

\section{Experimental Results}

\subsection{Baselines}

We first examine how \texttt{Phi-4-Multimodal} performs in the broader landscape of LLM-integrated ASR systems. 
The top portion of Table~\ref{tab:verbose_hybrid} compares \texttt{Phi-4-Multimodal} with the aggregated best results from prior LLM-based corrective ASR methods on the Open ASR Leaderboard \citep{chen2023hyporadise,lin-etal-2025-neko,wan2026speech}. Beam search with \texttt{Phi-4-Multimodal} achieves the best WER on 5 of the 7 benchmarks and outperforms the ensemble of prior corrective methods overall, which also incorporate beam search within their correction loops.
This performance is achieved with fewer parameters than existing LLM-based corrective approaches, highlighting the effectiveness of warm-initialized LLM-based ASR models and motivating further exploration of self-correction directly within \texttt{Phi-4-Multimodal}.


\subsection{Hybrid Search for Decoding}

The rows highlighted in green in Table~\ref{tab:verbose_hybrid} compare Hybrid Search ($\tau=\frac{1}{2}$) with standard decoding methods. 
Hybrid Search recovers 25\% of the improvement achieved by beam search over greedy search. 
This gain is substantially larger for named entities, where Hybrid Search recovers 96.4\% of the improvement achieved by beam search.
As Hybrid Search combines greedy search and beam search, its computational cost lies between those of the two decoding methods (Table~\ref{tab:hybrid_efficiency}). On average, Hybrid Search targets 10\% of tokens for beam expansion. With $B=5$, this gives an amortized beam size of 1.4 for Hybrid Search, corresponding to approximately \(1.4\times\) the greedy-decoding cost.
These results establish Hybrid Search as an efficient and effective decoding algorithm, especially for improving semantically important tokens.

How important is the LLM contribution within the Hybrid Search algorithm? To isolate its contribution, we keep the Hybrid Search algorithm unchanged but remove LLM rescoring by setting $\alpha=0$. This variant is denoted as ``- LLM Correction'' in Table~\ref{tab:verbose_hybrid}. 
While removing the LLM contribution achieves comparable WER, its named-entity improvement drops to 68.3\% of the beam-search gain (from 96.4\%).
These results suggest that the targeted regions identify uncertain tokens that benefit not only from additional search, but also from the linguistic information provided by the base LLM. 
Since our method is self-corrective, this result further suggests that speech-recognition adaptation may induce partial forgetting of semantic knowledge, motivating future work on more effective adaptation strategies for LLM-to-ASR-LLM training.


\subsection{Hybrid Search for Error Correction}

We combine Hybrid Search with beam search to improve decoding robustness. Rows highlighted in yellow in Table~\ref{tab:verbose_hybrid} show the results of Hybrid Search compared to Beam Search and naive LLM correction methods such as rescoring and late fusion.
We observe that Hybrid Search improves named-entity recognition over beam search while maintaining comparable overall WER, where $\tau=\frac{1}{2}$ bounds the more favorable targeted region among the two thresholds. 
Under this setting, NE-ER decreases from 18.29\% to 17.69\%. The largest gains occur on TED-LIUM and SPGISpeech, where the NE-ER relative improvements are 12.7\% and 8.8\% respectively.
Two-sided exact McNemar tests are shown in Table~\ref{tab:verbose_hybrid} below the corresponding results, confirming that the improvements are statistically significant for both NE-ER ($p=3.56\times10^{-38}$) and WER ($p=9.19\times10^{-5}$).

The results indicate that our targeted correction method reliably improves difficult predictions, particularly named entities, even when such cases are sparse, without degrading WER. 
Nevertheless, because named entities substantially affect semantic fidelity and downstream utility, these gains are particularly meaningful \citep{wan2025speechiq}.
We observe that most edits occur within multi-token words, highlighting a systematic issue in how \texttt{Phi-4-Multimodal} adapts to ASR.

\section{Analysis}

\subsection{Analysis of Correction Examples}
\label{sec:case_examples}
Manual inspection of the edits introduced by our proposed Hybrid Search reveals that most edits, whether correct or incorrect, involve semantically specific expressions, particularly named entities. These include person names (e.g., \textit{Kim Zmeskal}), organizations (e.g., \textit{Cedefop}), and brands (e.g., \textit{Telenor}). Foreign and less common lexical items also frequently appear among the edited tokens, such as \textit{señor}, \textit{Guanyin}, and \textit{Khan al-Ahmar}. Overall, Hybrid Search primarily modifies tokens for which linguistic or semantic context is informative, while hallucination-inducing edits are minimal. Representative examples are provided in Appendix~\ref{apx:case_examples}.


\subsection{Is Targeted Correction Necessary?}

As an ablation of targeted hybrid search, we include non-targeted baselines, including naive rescoring from beam search and direct late fusion. From Table~\ref{tab:verbose_hybrid}, it is evident that direct late fusion performs remarkably worse (WER 6.36 $\rightarrow$ 9.11), as the results frequently introduce hallucinated outputs from undesired LLM guidance. In contrast, naive rescoring still degrades on WER but can improve on NE-ER. 

\begin{figure}[t]
    \centering
    \includegraphics[width=0.9\columnwidth]{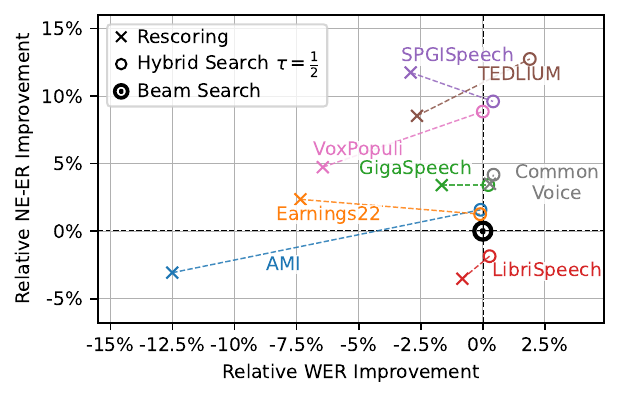}
    \caption{Relative WER and NE-ER improvements across test sets, measured against beam search. Dashed lines connect Rescoring and Hybrid Search results, showing that Hybrid Search substantially improves WER while maintaining NE-ER accuracy.}
    \label{fig:wer-ner}
\end{figure}

We visualize the comparison between rescoring, beam search, and Hybrid Search in Figure~\ref{fig:wer-ner}. Using beam search as the anchor, we plot the relative WER and NE-ER improvements of other experiments. Dashed lines connect each ``Rescoring'' result to its corresponding Hybrid Search result with $\tau=\frac{1}{2}$. 
The lines reveal that our method substantially improves WER performance while maintaining NE-ER compared to rescoring. 
These results show that targeted correction is necessary to improve NE-ER beyond beam search without sacrificing overall WER.

\subsection{Are Intermediate Interactions Necessary?}

We observe from~\ref{sec:hid-sem} and Section~\ref{sec:target-correction} that the distribution shift is largely driven by multi-token words, and the targeted words are heavily biased towards them.
Since multi-token words can be determined simply from the surface form of the output tokens, we discuss the necessity of including boundaries that are tied to hidden-state features. In Table~\ref{tab:verbose_hybrid}, we include a baseline with $\tau=0$ that targets only multi-token words and does not use any intermediate features. This setting fails to preserve the WER robustness of beam search, demonstrating that the proposed hidden-state interactions are essential for effective targeting.

\subsection{Complexity Analysis}
\label{sec:Complexity}

We elaborate on the efficiency of Hybrid Search for Decoding. The decoding latency of \texttt{Phi-4-Multimodal}, is bounded by the autoregressive decoding of $N$ steps. 
In our method, the ASR-LLM and its internal LLM can be executed in parallel, so the overall latency remains bounded by these same $N$ decoding steps.
The feature extraction cost is negligible compared with a model forward pass.
Specifically, computing $cos(\cdot)$ and $norm(\cdot)$ together requires approximately $10hL$ FLOPs, where $h=3072$ is the hidden size and $L=33$ is the total number of layers plus the embedding layer. 
This gives $10 \times 3072 \times 33 \approx 10^6$ FLOPs per decoding position, compared with a model forward pass whose computation is on the order of $10^9$ FLOPs.
The decoding complexity thus reduces to the selective beam-search cost, given by \(B_{\mathrm{eff}}=(1-\rho)+\rho B\). With a beam search expansion selection rate of \(\rho=0.1\) and beam size \(B=5\), this equals an amortized cost of \(1.4\times\) greedy decoding.

\section{Conclusions}

This paper introduces Hybrid Search, a self-correction algorithm for ASR-LLMs that combines targeted refinement with semantic-based correction. Using simple latent descriptors, including cosine similarity and relative norm between the ASR-LLM and its base-LLM counterpart, we identify clear distributional signals associated with semantically dependent tokens for \texttt{Phi-4-Multimodal} and \texttt{Granite-4.0-1B-Speech}. Experiments with \texttt{Phi-4-Multimodal} further show that our method improves NE-ER while maintaining competitive WER, without requiring external models or additional training. These results suggest that even for warm-initialized models, residual knowledge from the base LLM may remain exploitable at inference time.

\section*{Limitations}
While we validate the hidden-state observations on both \texttt{Phi-4-Multimodal} and \texttt{Granite-4.0-1B-Speech}, the end-to-end Hybrid Search experiments are conducted only on \texttt{Phi-4-Multimodal}. Therefore, the generalization of the full correction method across ASR-LLM architectures remains to be established. The exact feature distributions, layer aggregation, and decision thresholds may be model-dependent and require recalibration, and may not be available for closed-source systems. Although we observe generalization across the Open ASR Leaderboard and Common Voice, feature-space distributions can vary across models and domains. More adverse settings, such as singing ASR, may therefore require additional calibration or adaptation.

\bibliography{colm2026_conference}

\appendix

\section{Extended Related Work}
\noindent\textbf{LLM-based ASR.} Recent advances in speech foundation models have demonstrated strong generalization and multi-task capabilities, including multilingual ASR and speech translation, by scaling training data to hundreds of thousands or even millions of hours. A notable example is Whisper, an encoder–decoder Transformer trained on 680,000 hours of weakly labeled speech data that achieves robust performance across diverse domains and languages \cite{whisper}. Later versions further scale the training corpus to 5 million hours \cite{openai_whisper_large_v3}. Following this direction, the OWSM series~\citep{peng25c_interspeech} introduces fully transparent and reproducible speech foundation models trained on large-scale open speech datasets while achieving competitive performance with proprietary systems. Similarly, Canary \citep{canary} introduces a multilingual ASR and speech translation model based on a FastConformer encoder–decoder architecture that achieves competitive performance while being trained on significantly less data.

To further improve performance by incorporating the rich linguistic knowledge encoded in LLMs, recent work has explored integrating pretrained speech encoders with LLMs. These multimodal LLMs are trained on large-scale speech–text corpora, enabling them to perform various speech-related tasks while leveraging the semantic capabilities of LLMs. To preserve the pretrained knowledge of LLMs while maintaining training efficiency, these models often adopt parameter-efficient adaptation techniques such as Low-Rank Adaptation (LoRA). For example, Canary-Qwen-2.5B \cite{nvidia2025canaryqwen25b} combines the FastConformer speech encoder from the pretrained Canary model with a Qwen-2.5B LLM decoder and significantly improves ASR performance compared to the original Canary model. Similarly, Phi-4-Multimodal \citep{phi4_mm} integrates speech, vision, and text processing within a shared LLM-based architecture using a Mixture-of-LoRAs design, enabling strong ASR performance alongside multimodal reasoning capabilities. In this work, we adopt Phi-4-Multimodal as the ASR–LLM backbone and investigate how its internal semantic representations can be more effectively leveraged to improve ASR performance.

\noindent\textbf{Generative Error Correction and Agentic Setups.} To leverage the rich semantic and world knowledge encoded in LLMs, GER methods typically operate in a two-pass pipeline, where an LLM refines the N-best hypotheses produced by a separate ASR system. For example, HyPoradise \citep{chen2023hyporadise} establishes a benchmark for using LLMs to directly predict ground-truth transcriptions from $N$-best hypotheses, and Whispering LLaMA \citep{radhakrishnan2023whispering} integrates Whisper-based acoustic representations with an LLaMA decoder to incorporate both acoustic and linguistic information. More recently, NeKo \citep{lin-etal-2025-neko} explores cross-modal post-recognition correction using a mixture-of-experts architecture. However, these agentic approaches~\citep{yang2025spoken} rely on a sequential two-pass inference process, where the ASR model first generates candidate hypotheses and the LLM subsequently performs correction, introducing additional latency.

To address this limitation, recent work has explored single-pass, synchronous decoding frameworks that integrate LLMs more tightly into the search process. For instance, Generative Fusion Decoding \citep{hsu2025let} employs a byte-level shallow fusion to bridge mismatched token spaces between ASR and LLM decoders without requiring re-training. Similarly, SALSA \citep{mittal2024salsa} couples the hidden states of an ASR decoder to an LLM via a trained projection layer for synchronous hypothesis advancement. To address computational costs, delayed fusion \citep{hori2025delayed} applies LLM scores to ASR hypotheses only after a pruning stage, significantly reducing the number of required LLM inference calls. While these methods reduce the latency of two-pass GER pipelines, they still require parallel inference from two separate ASR and LLM models. In contrast, our proposed approach operates within a unified ASR-LLM architecture, directly leveraging the internal hidden states of the LLM backbone during a single forward pass to improve ASR performance with relatively small computational overhead.

\noindent\textbf{Analysis of LLM Hidden States.} With billions of parameters, LLMs learn internal representations that encode rich semantic and functional properties that are closely related to their generation behavior. Recent studies attempt to analyze these hidden representations to better understand how LLMs perform reasoning and decision-making. Specifically, the latent chain-of-thought (CoT) reasoning suggests that LLMs may perform intermediate inferential computations within their high-dimensional hidden states, even when no explicit natural-language reasoning is produced \citep{chen2025reasoning}. Moreover, the Chain-of-Embedding (CoE) framework \citep{coe} demonstrates that the sequence of hidden states across layers can be interpreted as a latent reasoning trajectory. By analyzing geometric properties of these trajectories, such as magnitude changes and angular differences between layer-wise embeddings, it becomes possible to estimate prediction correctness or detect semantic uncertainty without requiring ground-truth supervision.

\begin{algorithm}[t]
\caption{Hybrid Search}
\label{alg:hybrid-search}
\begin{algorithmic}[1]
\Require Audio input $A$, ASR-LLM $p_\theta$, 
         target detector $\mathcal{T}(\cdot)$, beam size $B$
\Ensure Final decoded sequence $y$

\State $y \leftarrow \emptyset$;
       $\mathcal{B} \leftarrow \{y\}$;
       $\text{phase} \leftarrow \textsc{Greedy}$;
       $K \leftarrow 1$
       \Comment{$K$: effective beam size}

\While{not end-of-sequence}

    \Statex \textbf{[Search]}
    \State \textbf{Advance one decoding step with $K$}
    \State $y \leftarrow$ best beam in $\mathcal{B}$;
           $y_t \leftarrow$ last token in $y$

    \Statex \textbf{[State transition]}
    \If{$\text{phase} = \textsc{Greedy}$}
        \If{$\mathcal{T}(y_t)$}
            \State $K \leftarrow B$;
                   $\mathcal{B} \leftarrow \{y\}^{B}$;
                   $\text{phase} \leftarrow \textsc{Beam}$
        \EndIf

    \ElsIf{$\text{phase} = \textsc{Beam}$}
        \If{$\exists\, y' \in \bigcap_{b \in \mathcal{B}} b
              \;\text{s.t.}\; \neg\mathcal{T}(y')$}
            \State Roll back each beam to the position ending at $y'$
            \State $y \leftarrow$ best rolled-back beam in $\mathcal{B}$;
                   $\mathcal{B} \leftarrow \{y\}$;
                   $K \leftarrow 1$;
                   $\text{phase} \leftarrow \textsc{Greedy}$
        \EndIf
    \EndIf

\EndWhile

\State \Return $y$
\end{algorithmic}
\end{algorithm}

\section{Detailed Experimental Setting}

\subsection{Late Layer Feature Discovery}
\label{apx:late_feat}
In Figure~\ref{fig:gaussian}, we isolate the $\mathbf{AngleFeat}$ and $\mathbf{MagFeat}$ from the hidden state before the final layer (layer 31). For named-entity tokens, the overall distribution shifts toward greater deviation (lower cosine, higher norm). In particular, this phenomenon is most pronounced for starting tokens of multi-token words.  We attribute this behavior to uncertainty in the ASR-LLM's final token selection among acoustically similar candidates. Multi-token named entities may contain spelling variations that require longer context to resolve, making an immediate decision uncertain (``Carol -yn'' vs. ``Kar -ol -yn''). Constructing an embedding space that supports a more balanced probability mixture over competing candidates requires a larger departure from the original LLM embedding space, as that space is primarily organized around semantic relationships and, therefore, tends to underemphasize acoustic similarity.

\subsection{Hybrid Search Algorithm}
\label{apx:hybrid_algo}

Algorithm~\ref{alg:hybrid-search} presents the Hybrid Search Algorithm. The algorithm switches between greedy search states and beam search states. We use the term ``Advance one decoding step'' to convey both stepping forward in beam search and greedy search depending on the current state of the algorithm.

\begin{figure}[t]
    \centering
    \includegraphics[width=\columnwidth]{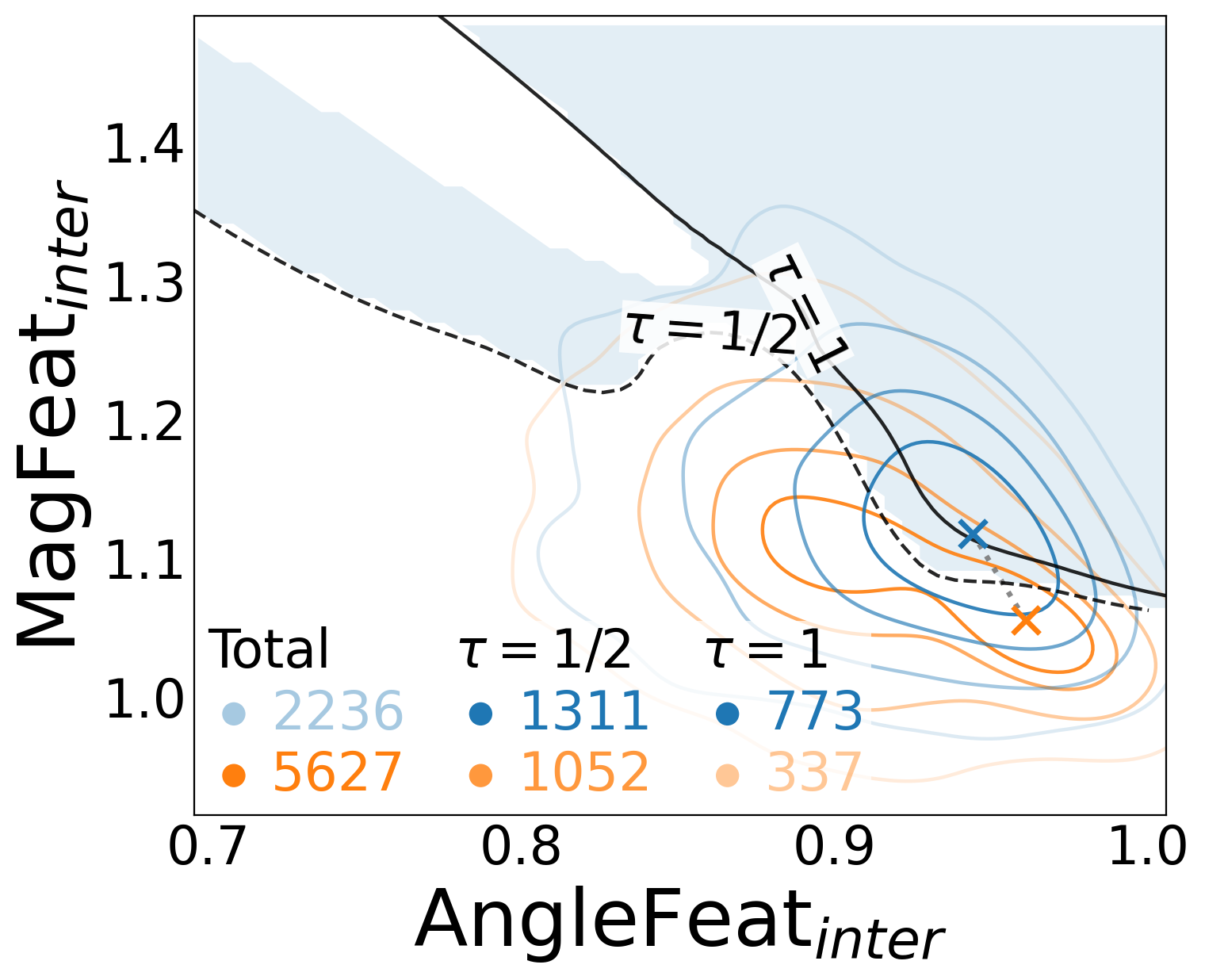}
    \caption{Named-Entity Distribution Shifts of Granite-4.0-1B-Speech}
    \label{fig:granite_gaussian}
\end{figure}

\section{Additional Results and Analysis}

\subsection{Generalization of targeted approaches}
\label{apx:granite}
We show similar distribution shifts in $\mathrm{AngleFeat}_{inter}$ and $\mathrm{MagFeat}_{inter}$ for \texttt{Granite-4.0-1B-Speech} in Figure~\ref{fig:granite_gaussian}.

\subsection{Generalization of targeted region}
\begin{figure*}[t]
\centering

\begin{tabular}{ccc}

\texttt{YODAS} & & \\[2pt]
\includegraphics[width=0.31\textwidth]{figs2/EXP10_-_nonner_-_ner_-_gaussian-with-decision-no-space-only.png} & &\\[6pt]

{\Large \textbf{$\Downarrow$}} & & \\[2pt]

\texttt{AMI} & \texttt{Common Voice} & \texttt{Earnings22} \\[2pt]
\includegraphics[width=0.31\textwidth]{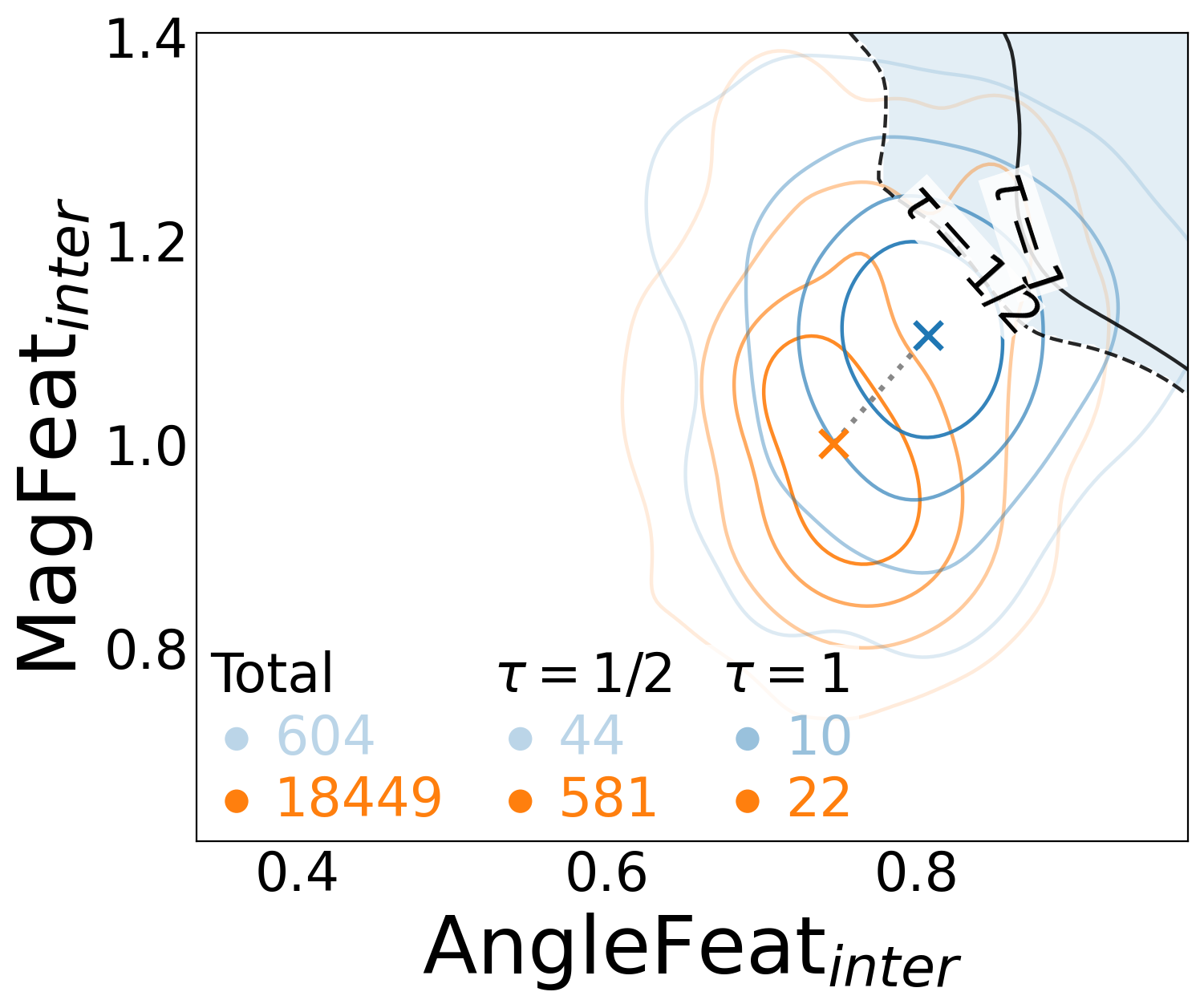} &
\includegraphics[width=0.31\textwidth]{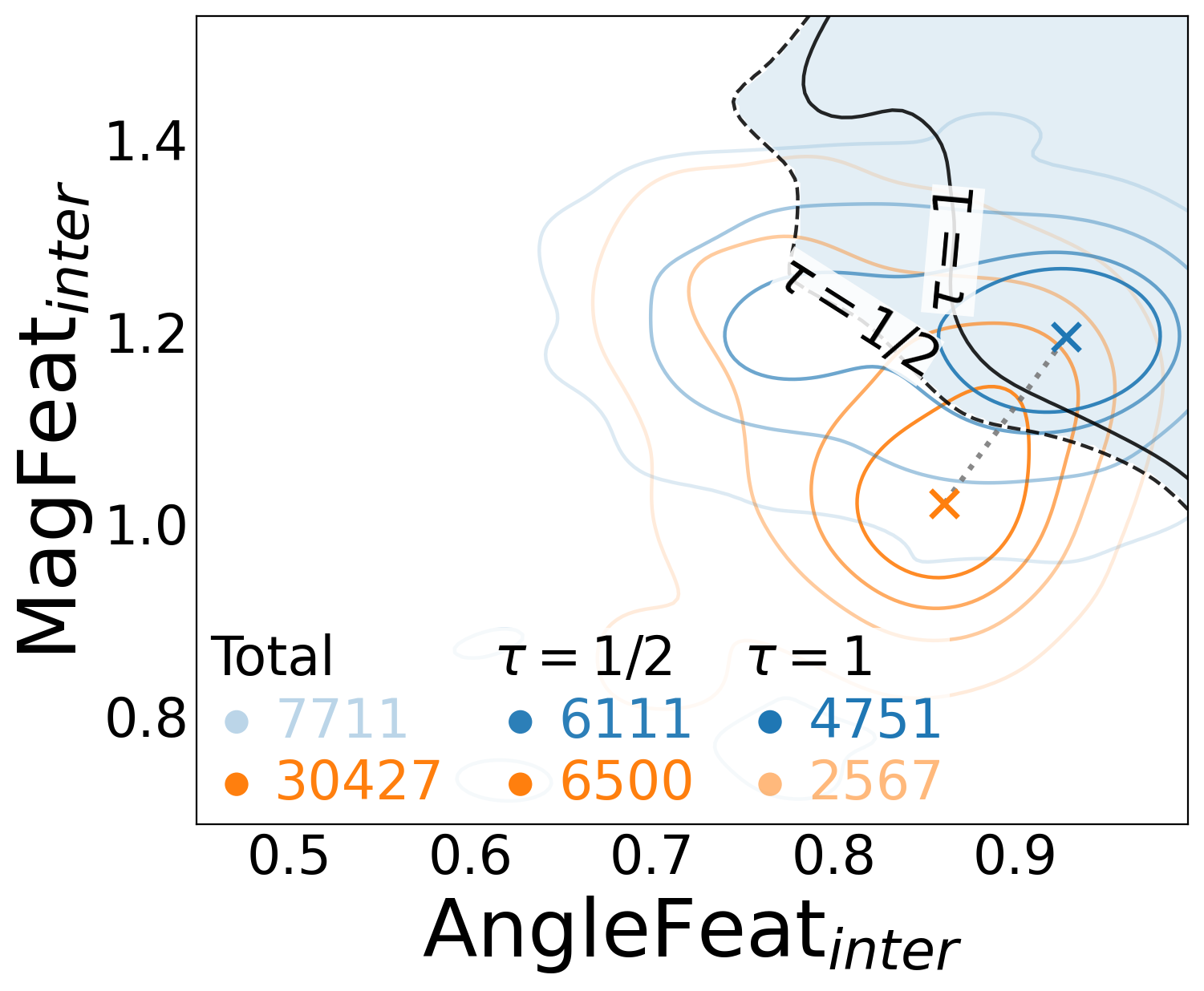} &
\includegraphics[width=0.31\textwidth]{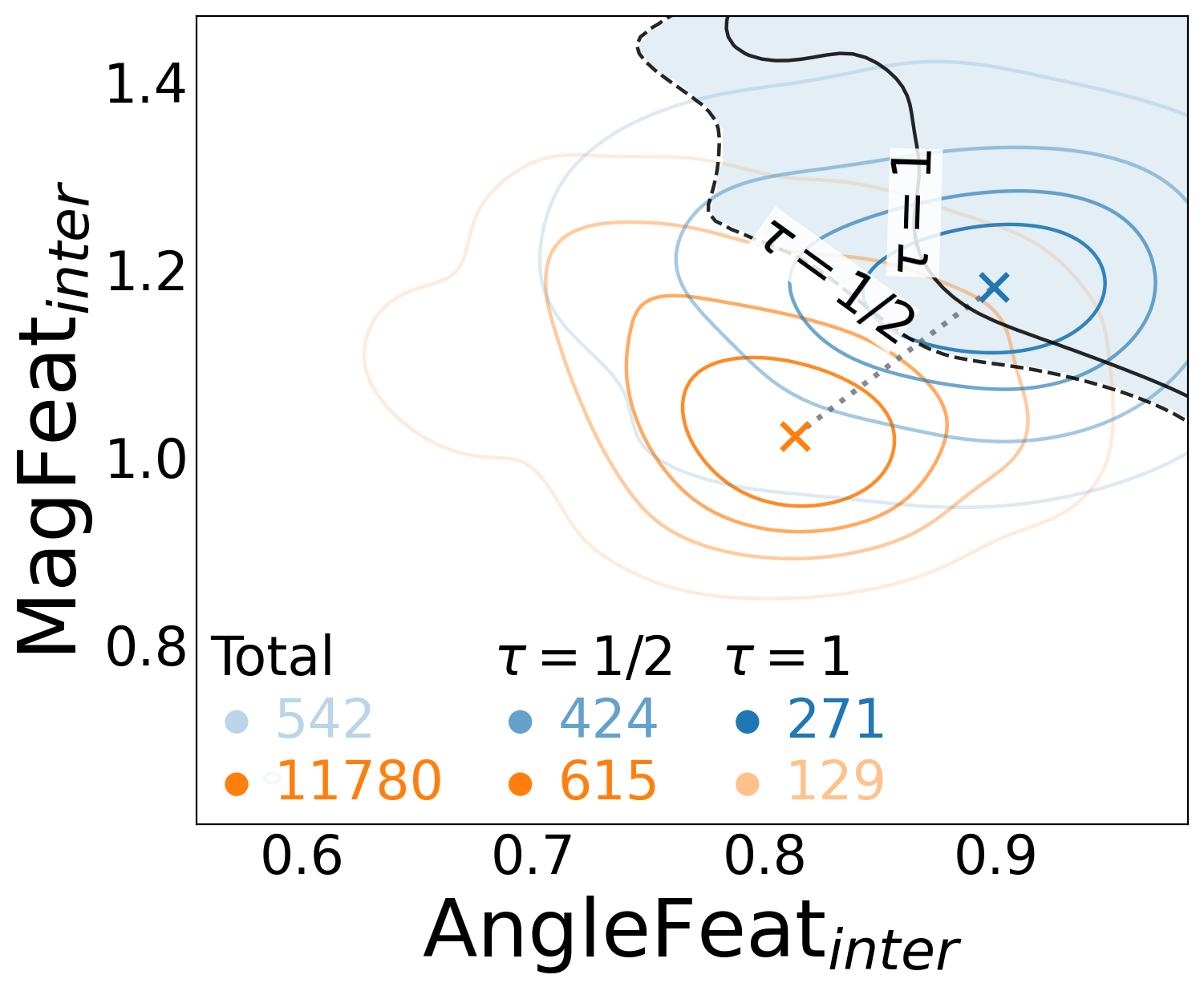} \\[6pt]

\texttt{GigaSpeech} & \texttt{LibriSpeech} & \texttt{SPGISpeech} \\[2pt]
\includegraphics[width=0.31\textwidth]{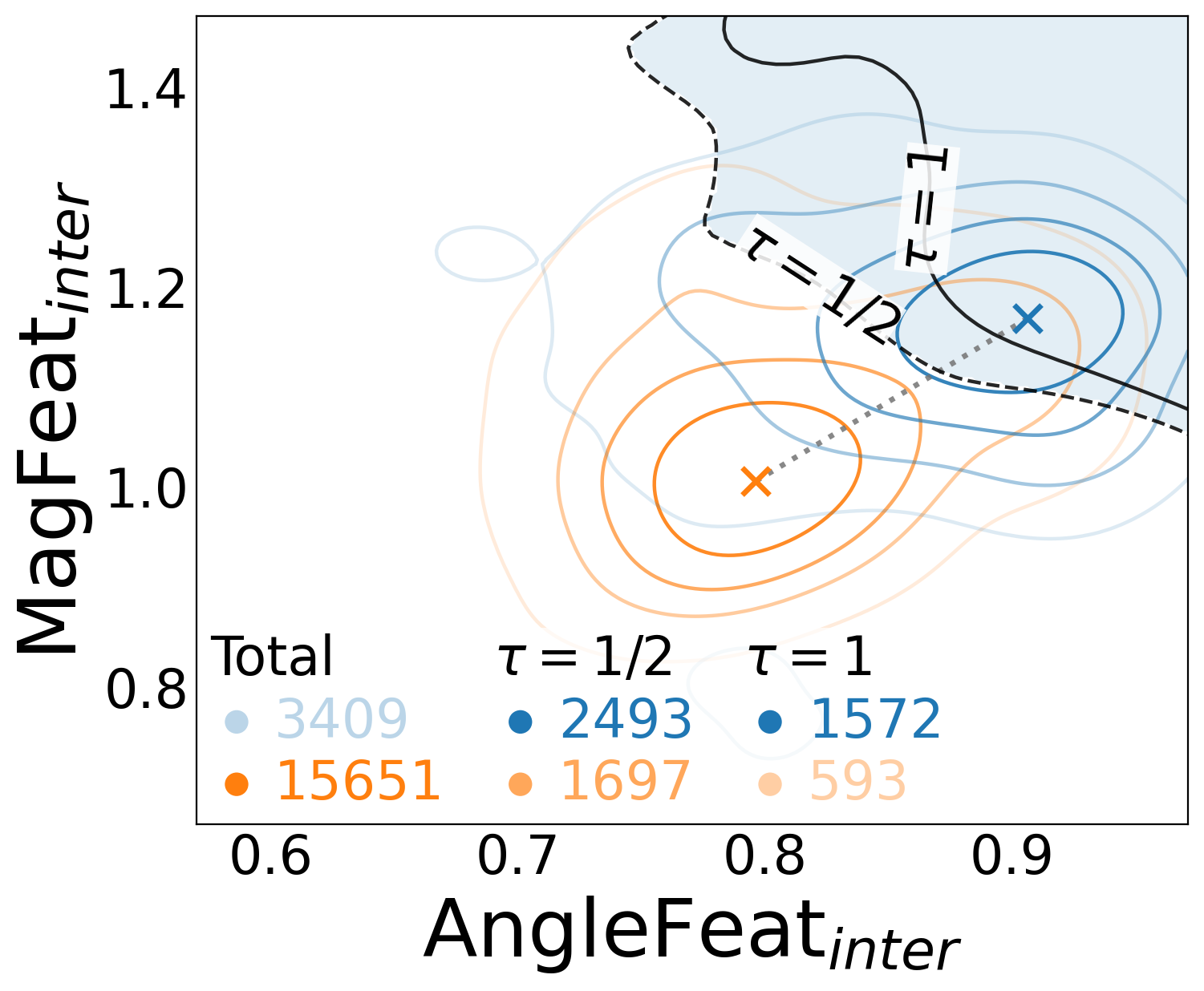} &
\includegraphics[width=0.31\textwidth]{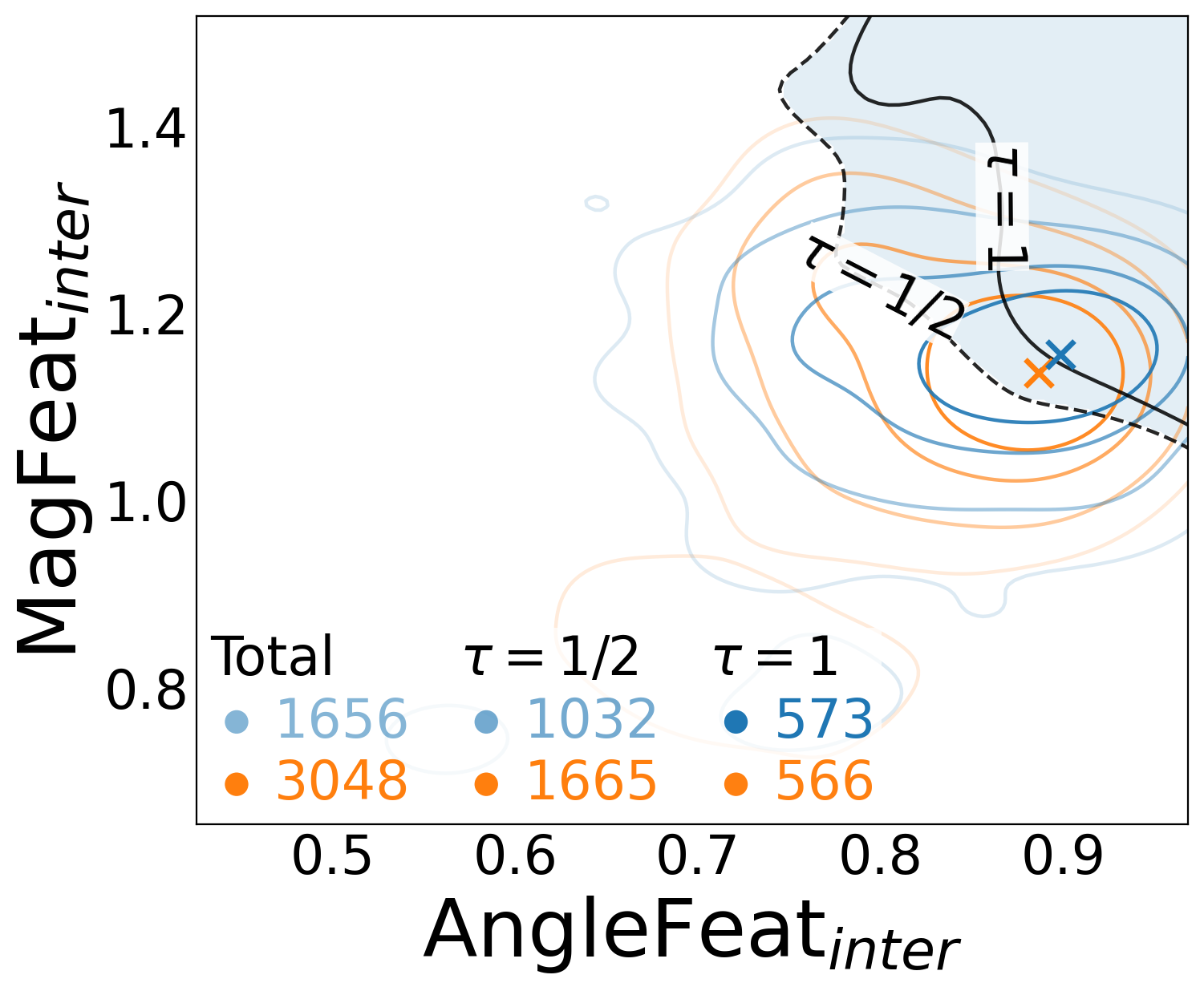} &
\includegraphics[width=0.31\textwidth]{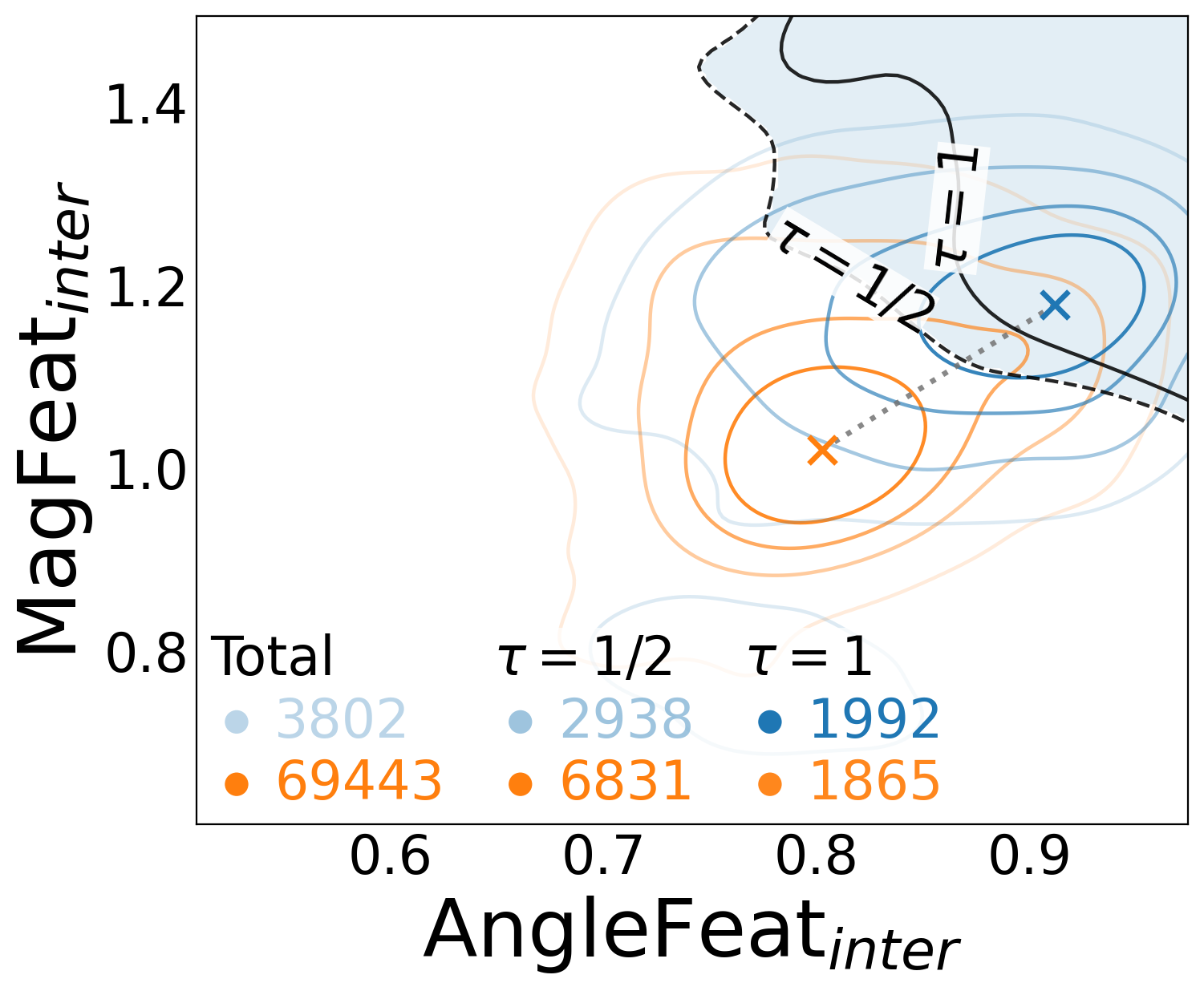} \\[6pt]

\texttt{TED-LIUM} & \texttt{VoxPopuli} & \\[2pt]
\includegraphics[width=0.31\textwidth]{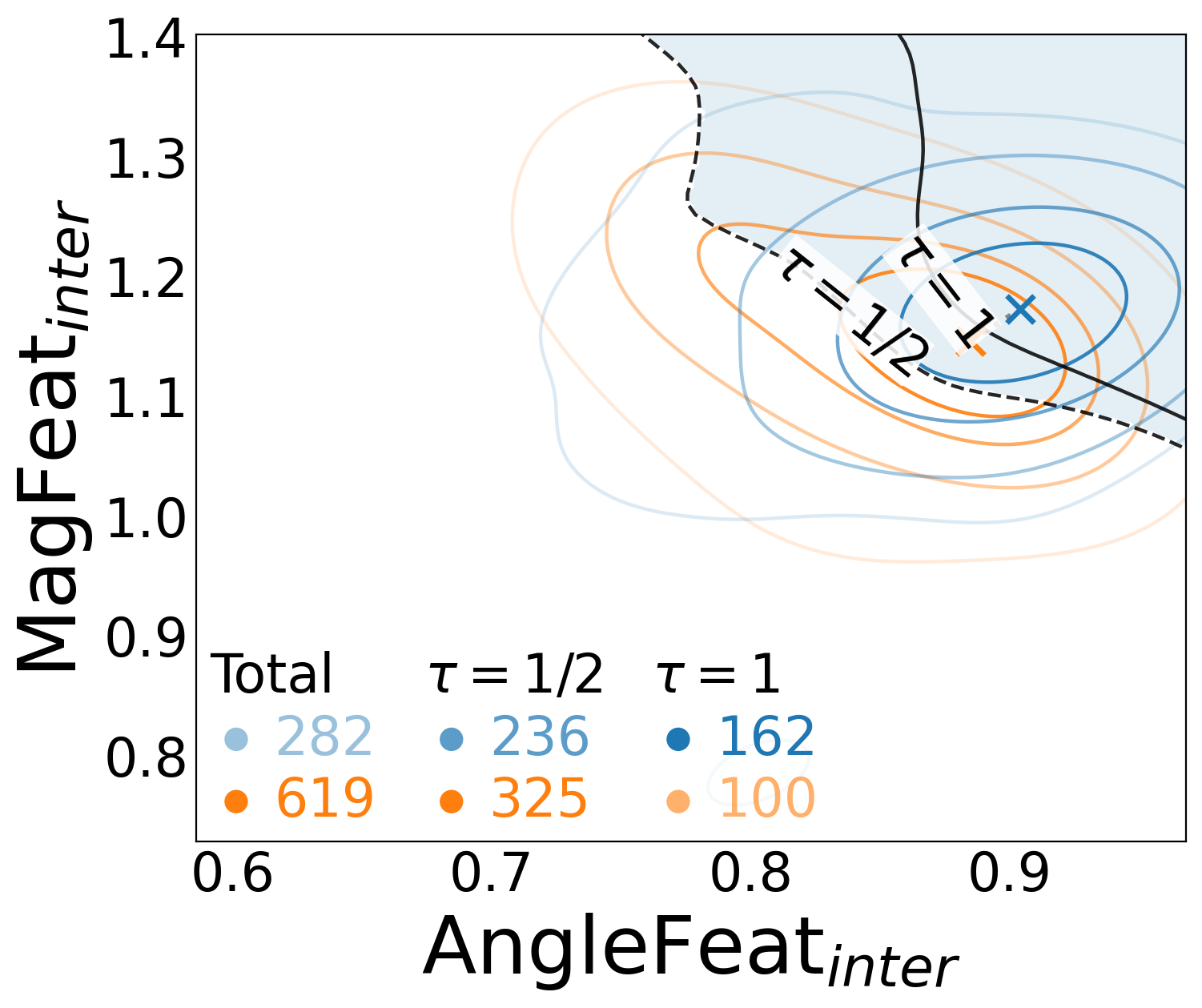} &
\includegraphics[width=0.31\textwidth]{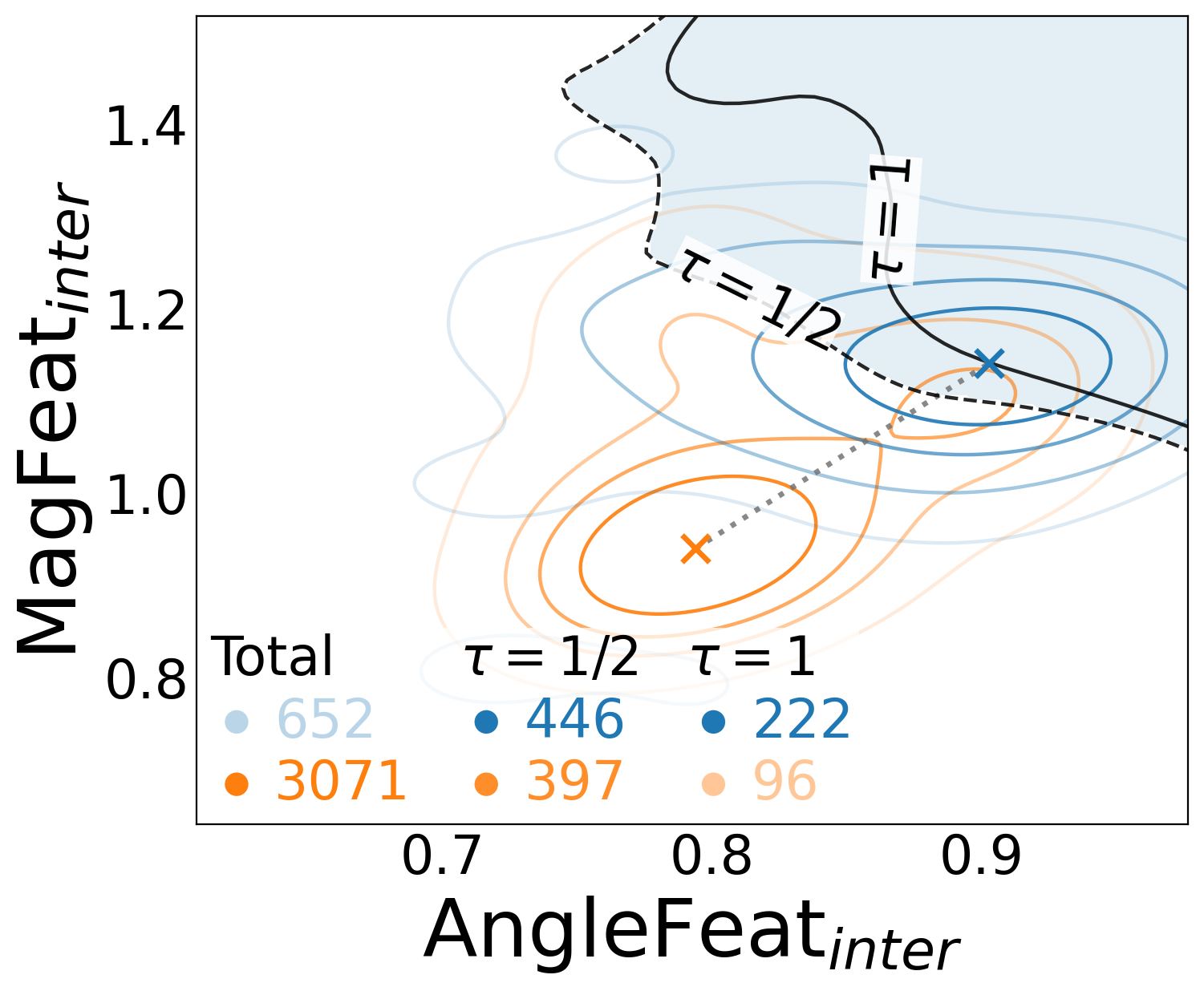} &
\\[6pt]
\end{tabular}

\caption{Generalization of the targeted region derived from the \texttt{YODAS} dataset. Most datasets exhibit a similar distribution shift. Under the targeted region selection derived from the \texttt{YODAS} dataset, named entities are effectively concentrated.}
\label{fig:targeted_regions}
\end{figure*}

In Figure~\ref{fig:targeted_regions}, we highlight the targeted region derived from \texttt{YODAS}, and show how it generalizes to other datasets. Applying the region to other datasets in the Open ASR Leaderboard, we observe consistent concentration effects in named-entity tokens. 

\subsection{Case Examples of Correction}
\label{apx:case_examples}
We list the first few chronological instances from each of the datasets in which Hybrid Search introduces edit, and separate into positive and negative edit categories. Significant edits mostly involve named entities, while changes to acoustically dominant tokens that risk introducing hallucinations are relatively uncommon.

\begin{tcolorbox}[width=\columnwidth, enhanced, breakable]
\small
\setlength{\parindent}{0pt}

\textbf{[Common Voice]}

Beam Search: assemblyman james silva and former state senator \textbf{ventran} boy scouts of america

Hybrid Search $\tau=\frac{1}{2}$: assemblyman james silva and former state senator \textbf{van tran} boy scouts of america

Gold: assemblyman james silva and former state senator \textbf{van tran} boy scouts of america

\medskip
Beam Search: in this performance she sang with example \textbf{sigfried} jerusalem and john tomlinson

Hybrid Search $\tau=\frac{1}{2}$: in this performance she sang with example \textbf{siegfried} jerusalem and john tomlinson

Gold: in this performance she sang with example \textbf{siegfried} jerusalem and john tomlinson

\medskip
Beam Search: it starred hawks jack dee doon mackichan ben miller alistair \textbf{mcgregor} and charlotte page

Hybrid Search $\tau=\frac{1}{2}$: it starred hawks jack dee doon mackichan ben miller alistair \textbf{mcgowan} and charlotte page

Gold: it starred hawks jack dee doon mackichan ben miller alistair \textbf{mcgowan} and charlotte page
\end{tcolorbox}

\begin{tcolorbox}[width=\columnwidth, enhanced, breakable]
\small
\setlength{\parindent}{0pt}

\textbf{[AMI]}

\begin{center}
\colorbox{green!12}{\strut\textbf{\textcolor{green!50!black}{Correct Hybrid edits}}}
\end{center}

Beam Search: and i think we should show this prototype to people from various age and \textcolor{red}{\textbf{socioeconomic}} groups and see about any fine tuning that maybe little things that we have not thought of

Hybrid Search $\tau=\frac{1}{2}$: and i think we should show this prototype to people from various age and \textcolor{green!50!black}{\textbf{socio economic}} groups and see about any fine tuning that maybe little things that we have not thought of

Gold: and i think we should show this prototype to people from various age and \textbf{socio economic} groups and see about any fine tuning that maybe little things we have not thought of

\medskip

Beam Search: the design should minimize \textcolor{red}{\textbf{our}} s i and be easy to locate and we were still a slightly ambivalent as to whether to use voice recognition there

Hybrid Search $\tau=\frac{1}{2}$: the design should minimize \textcolor{green!50!black}{\textbf{r}} s i and be easy to locate and we were still a slightly ambivalent as to whether to use voice recognition there

Gold: the design should minimize \textbf{r} s i and be easy to locate and we were still slightly ambivalent as to whether to use voice recognition there

\medskip

Beam Search: it does does an n x t search for the string i \textcolor{red}{\textbf{dunno}} transcript or text or whatever

Hybrid Search $\tau=\frac{1}{2}$: it does does an n x t search for the string i \textcolor{green!50!black}{\textbf{do not know}} transcript or text or whatever

Gold: it does does an n x t search for the string i \textbf{do not know} transcript or text or whatever

\medskip
\hrule
\medskip

\begin{center}
\colorbox{red!10}{\strut\textbf{\textcolor{red!70!black}{Incorrect Hybrid edits}}}
\end{center}

Beam Search: now these features includes the s signal emitting signal it is the m led or l e d the \textcolor{green!50!black}{\textbf{infrared}}

Hybrid Search $\tau=\frac{1}{2}$: now these features includes the s signal emitting signal it is the m led or l e d the \textcolor{red}{\textbf{infra red}}

Gold: now these features includes the s signal emitting signal it is the led or l e d the \textbf{infrared}

\medskip

Beam Search: actually for for an \textcolor{green!50!black}{\textbf{n}} text area whatever it is they they define some ha handy highlights on that

Hybrid Search $\tau=\frac{1}{2}$: actually for for an \textcolor{red}{\textbf{in}} text area whatever it is they they define some ha handy highlights on that

Gold: actually for for an \textbf{n} text area whatever it is they they defined some ha handy highlights on that

\medskip

Beam Search: cause in our earlier market research if you would allow me to go to the \textcolor{green!50!black}{\textbf{flat board}} smart board

Hybrid Search $\tau=\frac{1}{2}$: cause in our earlier market research if you would allow me to go to the \textcolor{red}{\textbf{whiteboard}} smart board

Gold: cause in our earlier market research if you would allow me to go to the \textbf{flat board} smartboard
\end{tcolorbox}

\begin{tcolorbox}[width=\columnwidth, enhanced, breakable]
\small
\setlength{\parindent}{0pt}

\textbf{[Common Voice]}

\begin{center}
\colorbox{green!12}{\strut\textbf{\textcolor{green!50!black}{Correct Hybrid edits}}}
\end{center}

Beam Search: it was the time of day when all of \textcolor{red}{\textbf{spence}} slept during the summer

Hybrid Search $\tau=\frac{1}{2}$: it was the time of day when all of \textcolor{green!50!black}{\textbf{spain}} slept during the summer

Gold: it was the time of day when all of \textbf{spain} slept during the summer

\medskip

Beam Search: due to her symbolization of compassion in east asia \textcolor{red}{\textbf{guan yin}} is associated with vegetarianism

Hybrid Search $\tau=\frac{1}{2}$: due to her symbolization of compassion in east asia \textcolor{green!50!black}{\textbf{guanyin}} is associated with vegetarianism

Gold: due to her symbolization of compassion in east asia \textbf{guanyin} is associated with vegetarianism

\medskip

Beam Search: from the vietnam era and to present \textcolor{red}{\textbf{purposes}} built ceremonial knives evolved

Hybrid Search $\tau=\frac{1}{2}$: from the vietnam era and to present \textcolor{green!50!black}{\textbf{purpose}} built ceremonial knives evolved

Gold: from the vietnam era and to present \textbf{purpose} built survival knives evolved

\medskip
\hrule
\medskip

\begin{center}
\colorbox{red!10}{\strut\textbf{\textcolor{red!70!black}{Incorrect Hybrid edits}}}
\end{center}

Beam Search: the ivy \textcolor{green!50!black}{\textbf{climbed}} up the building and wrapped itself around the chimney

Hybrid Search $\tau=\frac{1}{2}$: the ivy \textcolor{red}{\textbf{clambered}} up the building and wrapped itself around the chimney

Gold: the ivy \textbf{climbed} up the building and wrapped itself around the chimney

\medskip

Beam Search: 2 theories exist as to how \textcolor{green!50!black}{\textbf{loco}} acquired its name

Hybrid Search $\tau=\frac{1}{2}$: 2 theories exist as to how \textcolor{red}{\textbf{locarno}} acquired its name

Gold: 2 theories exist as to how \textbf{loco} acquired its name

\medskip

Beam Search: after leaving ford shinoda and knudsen co created \textcolor{green!50!black}{\textbf{rectrans}} which built recreational vehicles

Hybrid Search $\tau=\frac{1}{2}$: after leaving ford shinoda and knudsen co created \textcolor{red}{\textbf{rekchamps}} which built recreational vehicles

Gold: after leaving ford shinoda and knudsen co created \textbf{rectrans} which built recreational vehicles
\end{tcolorbox}

\begin{tcolorbox}[width=\columnwidth, enhanced, breakable]
\small
\setlength{\parindent}{0pt}

\textbf{[Earnings-22]}

\begin{center}
\colorbox{green!12}{\strut\textbf{\textcolor{green!50!black}{Correct Hybrid edits}}}
\end{center}

Beam Search: if we look into take this forward then what you can expect is that we you know in the in 22 we will be able as we recover from covid although we still have some weakness in you know the end of january and in january as we speak we will be able to get to a better sustainable commercial rate but not at the level of q 4 not the high level of q 4 because that is a seasonality and not at the low levels we have seen earlier in in 21 where we were very much focused on focused on ebitda in the midst of covid

Hybrid Search $\tau=\frac{1}{2}$: if we look into take this forward then what you can expect is that we you know in the \textcolor{green!50!black}{\textbf{in}} in 22 we will be able as we recover from covid although we still have some weakness in you know the end of january and in january as we speak we will be able to get to a better sustainable commercial rate but not at the level of q 4 not the high level of q 4 because that is a seasonality and not at the low levels we have seen earlier in in 21 where we were very much focused on focused on ebitda in the midst of covid

Gold: if we look into take this forward then what you can expect is that we you know \textbf{in} in 22 we will be able as we recover from covid although we still have some weakness in you know the end of january and in january as we speak we will be able to get to a better sustainable commercial rate but not at the level of q 4 not the high level of q 4 because that is a seasonality and not at the low levels we have seen earlier in in 2021 where we were very much focused on focused on ebitda in the midst of covid

\medskip

Beam Search: good morning i will take the 1st part of your question in relation to what part of the growth on health and nutrition was sort of underlying growth versus a one \textcolor{red}{\textbf{offs}} and then i will pass it on to lisa in relation to the infant formula in market potential

Hybrid Search $\tau=\frac{1}{2}$: good morning i will take the 1st part of your question in relation to what part of the growth on health and nutrition was sort of underlying growth versus a one \textcolor{green!50!black}{\textbf{off}} and then i will pass it on to lisa in relation to the infant formula in market potential

Gold: i will take the 1st part of your question in relation to what part of the growth on health and nutrition was sort of underlying growth vs a one \textbf{off} and then i will pass it on to lisa in relation to the infant formula in market potential

\medskip

Beam Search: on the marketplace i focus on sweden and in regards to everything spoken about before the positioning in networks and your 5 g position and the jv with \textcolor{red}{\textbf{..}}

Hybrid Search $\tau=\frac{1}{2}$: on the marketplace i focus on sweden and in regards to everything spoken about before the positioning in networks and your 5 g position and the jv with \textcolor{green!50!black}{\textbf{telenor}}

Gold: on the marketplace i focus on sweden and in regard to everything spoken about before the positioning in networks and your 5 g position and the jv with \textbf{telenor}

\medskip
\hrule
\medskip

\begin{center}
\colorbox{red!10}{\strut\textbf{\textcolor{red!70!black}{Incorrect Hybrid edits}}}
\end{center}

Beam Search: so so given that in q 2 and q 4 last year \textcolor{green!50!black}{\textbf{hmos}} were 6 and 7000000 in revenue respectively i just wonder if sort of the underlying run rate for \textcolor{green!50!black}{\textbf{hmos}} was in fact weak given that you you you have this big product launch from from abbott

Hybrid Search $\tau=\frac{1}{2}$: so so given that in q 2 and q 4 last year \textcolor{red}{\textbf{hmo is}} were 6 and 7000000 in revenue respectively i just wonder if sort of the underlying run rate for \textcolor{red}{\textbf{hmo is}} was in fact weak given that you you you have this big product launch from from abbott

Gold: so so given that in q 2 and q 4 last year \textbf{hmos} were 6 and 7000000 in revenue respectively i just wonder if sort of the underlying run rate for \textbf{hmos} was in fact weak given that you you you have this big product launch fro from abbott

\medskip

Beam Search: excluding extraordinary charges comparable adjusted ebitda near breakeven reaching a loss of \textcolor{green!50!black}{\textbf{3600000}} improving from losses of 10500000 in the prior quarter and nearly 17000000 in 3rd quarter 2020

Hybrid Search $\tau=\frac{1}{2}$: excluding extraordinary charges comparable adjusted ebitda near breakeven reaching a loss of \textcolor{red}{\textbf{\$3600000}} improving from losses of 10500000 in the prior quarter and nearly 17000000 in 3rd quarter 2020

Gold: excluding extraordinary charges comparable adjusted ebitda neared break even reaching a loss of \textbf{3600000} improving from losses of 10 and a half 1000000 in the 3rd quarter and nearly 17000000 in 3rd quarter 2020

\medskip

Beam Search: while this resulted in a lower use of cash of close to 10000000 in 2nd quarter \textcolor{green!50!black}{\textbf{21}} with better travel conditions in this past quarter we saw a pickup in demand with customers redeeming travel vouchers

Hybrid Search $\tau=\frac{1}{2}$: while this resulted in a lower use of cash of close to 10000000 in 2nd quarter \textcolor{red}{\textbf{2021}} with better travel conditions in this past quarter we saw a pickup in demand with customers redeeming travel vouchers

Gold: while this resulted in a lower use of cash of close to 10000000 in 2nd quarter \textbf{21} with better travel conditions in this past quarter we saw a pick up in demand with customers redeeming travel vouchers
\end{tcolorbox}

\begin{tcolorbox}[width=\columnwidth, enhanced, breakable]
\small
\setlength{\parindent}{0pt}

\textbf{[GigaSpeech]}

\begin{center}
\colorbox{green!12}{\strut\textbf{\textcolor{green!50!black}{Correct Hybrid edits}}}
\end{center}

Beam Search: this is heavy metals from 30 for 30 podcasts i am alyssa rohnick gymnastics has been part of my life in some form for as long as i can remember from watching marylou on tv at the 1984 olympics when i was 7 to later being a teenage gymnast obsessing over kim \textcolor{red}{\textbf{zmesquel}} is tumbling passes

Hybrid Search $\tau=\frac{1}{2}$: this is heavy metals from 30 for 30 podcasts i am alyssa rohnick gymnastics has been part of my life in some form for as long as i can remember from watching marylou on tv at the 1984 olympics when i was 7 to later being a teenage gymnast obsessing over kim \textcolor{green!50!black}{\textbf{zmeskal}} is tumbling passes

Gold: this is heavy medals from 30 for 30 podcasts i am alyssa roenigk gymnastics has been part of my life in some form for as long as i can remember from watching mary lou on t v at the 1984 olympics when i was 7 too later being a teenage gymnast obsessing over kim \textbf{zmeskal} is tumbling passes

\medskip

Beam Search: when i covered my 1st summer olympic games in beijing in 2008 china was accused of having an underage gymnast on the team and i was writing about it i will never forget bela \textcolor{red}{\textbf{karoly}} shouting into my digital recorder about how the chinese coaches had just stolen his playbook

Hybrid Search $\tau=\frac{1}{2}$: when i covered my 1st summer olympic games in beijing in 2008 china was accused of having an underage gymnast on the team and i was writing about it i will never forget bela \textcolor{green!50!black}{\textbf{karolyi}} shouting into my digital recorder about how the chinese coaches had just stolen his playbook

Gold: when i covered my 1st summer olympic games in beijing in 2008 china was accused of having an underaged gymnast on the team and i was writing about it i will never forget bela \textbf{karolyi} shouting into my digital recorder about how the chinese coaches had just stolen his playbook

\medskip

Beam Search: and of course most of these critics were pop writers who were only aware of jeannie c \textcolor{red}{\textbf{reilly}} because of the pop success of harbor valley pta and whose only other frame of reference for country music in the 60 s and 70 s was probably johnny cash

Hybrid Search $\tau=\frac{1}{2}$: and of course most of these critics were pop writers who were only aware of jeannie c \textcolor{green!50!black}{\textbf{riley}} because of the pop success of harbor valley pta and whose only other frame of reference for country music in the 60 s and 70 s was probably johnny cash

Gold: and of course most of these critics were pop writers who were only aware of jeannie c \textbf{riley} because of the pop success of harper valley p t a and whose only other frame of reference for country music in the 60s and 70s was probably johnny cash

\medskip
\hrule
\medskip

\begin{center}
\colorbox{red!10}{\strut\textbf{\textcolor{red!70!black}{Incorrect Hybrid edits}}}
\end{center}

Beam Search: it is 10439493 people watching now simply better showed up and all it is like this is a red carpet event we got all the stars here tonight kg tropical simply better friday fish facts bob \textcolor{green!50!black}{\textbf{caylor}}

Hybrid Search $\tau=\frac{1}{2}$: it is 10439493 people watching now simply better showed up and all it is like this is a red carpet event we got all the stars here tonight kg tropical simply better friday fish facts bob \textcolor{red}{\textbf{keller}}

Gold: it is 10439493 people watching now simply brtta showed up and all that it is like this is a red carpet of me we got all the stars here tonight k g tropical simply betta friday fish facts bob \textbf{caylor}

\medskip

Beam Search: so that meant that tweens in the philippines or romania or macau could all rock a boxy \textcolor{green!50!black}{\textbf{shirred}} flounce cuff top or leopard drawstring lounge shorts for less than the price of a movie ticket

Hybrid Search $\tau=\frac{1}{2}$: so that meant that tweens in the philippines or romania or macau could all rock a boxy \textcolor{red}{\textbf{shouldered}} flounce cuff top or leopard drawstring lounge shorts for less than the price of a movie ticket

Gold: so that meant that tweens in the philippines or romania or macao could all rock a boxy \textbf{shirred} flounce cuff top or leopard drawstring lounge shorts for less than the price of a movie ticket

\medskip

Beam Search: blockbuster and \textcolor{green!50!black}{\textbf{walmart}} however were a bit late to the party hastings admitted that if they had started 2 years earlier they probably would have won on an end of year conference call hastings declared blockbuster has thrown everything but the kitchen sink at us

Hybrid Search $\tau=\frac{1}{2}$: blockbuster and \textcolor{red}{\textbf{wal mart}} however were a bit late to the party hastings admitted that if they had started 2 years earlier they probably would have won on an end of year conference call hastings declared blockbuster has thrown everything but the kitchen sink at us

Gold: blockbuster and \textbf{walmart} however were a bit late to the party hastings admitted that if they had started 2 years earlier they probably would have won on an end of year conference call hastings declared blockbuster has thrown everything but the kitchen sink at us
\end{tcolorbox}

\begin{tcolorbox}[width=\columnwidth, enhanced, breakable]
\small
\setlength{\parindent}{0pt}

\textbf{[LibriSpeech]}

\begin{center}
\colorbox{green!12}{\strut\textbf{\textcolor{green!50!black}{Correct Hybrid edits}}}
\end{center}

Beam Search: the egyptian obeyed and his master crossed the wide space strewn with sand and approached the stage which had been erected for the \textcolor{red}{\textbf{feudal}} performances even had his eyes retained the power of sight his blood was coursing so wildly through his veins that he might perhaps have been unable to distinguish the statues around him and the 1000s of spectators who crowded closely together richly garlanded their cheeks glowing with enthusiasm surrounded the arena hermon

Hybrid Search $\tau=\frac{1}{2}$: the egyptian obeyed and his master crossed the wide space strewn with sand and approached the stage which had been erected for the \textcolor{green!50!black}{\textbf{festal}} performances even had his eyes retained the power of sight his blood was coursing so wildly through his veins that he might perhaps have been unable to distinguish the statues around him and the 1000s of spectators who crowded closely together richly garlanded their cheeks glowing with enthusiasm surrounded the arena hermon

Gold: the egyptian obeyed and his master crossed the wide space strewn with sand and approached the stage which had been erected for the \textbf{festal} performances even had his eyes retained the power of sight his blood was coursing so wildly through his veins that he might perhaps have been unable to distinguish the statues around him and the 1000s of spectators who crowded closely together richly garlanded their cheeks glowing with enthusiasm surrounded the arena hermon

\medskip

Beam Search: sandro got up with pain enough in his bones and went after the innkeeper in the dark and meeting the officer who was looking to see what had become of his enemy he said to him \textcolor{red}{\textbf{seor}} whoever you are do us the favor and kindness to give us a little rosemary oil salt and wine for it is water to cure one of our best knights errant on earth who lies on yonder bed wounded by the hands of the enchanted moor that is in this inn

Hybrid Search $\tau=\frac{1}{2}$: sandro got up with pain enough in his bones and went after the innkeeper in the dark and meeting the officer who was looking to see what had become of his enemy he said to him \textcolor{green!50!black}{\textbf{senor}} whoever you are do us the favor and kindness to give us a little rosemary oil salt and wine for it is water to cure one of our best knights errant on earth who lies on yonder bed wounded by the hands of the enchanted moor that is in this inn

Gold: sancho got up with pain enough in his bones and went after the innkeeper in the dark and meeting the officer who was looking to see what had become of his enemy he said to him \textbf{senor} whoever you are do us the favor and kindness to give us a little rosemary oil salt and wine for it is wanted to cure one of the best knights errant on earth who lies on yonder bed wounded by the hands of the enchanted moor that is in this inn

\medskip

Beam Search: true an interesting conversation still had power to charm him but often during its continuance the full consciousness of his misfortune forced itself upon his mind for the majority of the subjects discussed by the artists came to them through the medium of sight and referred to new creations of architecture sculpture and painting from whose enjoyment his blindness \textcolor{red}{\textbf{debared}} him

Hybrid Search $\tau=\frac{1}{2}$: true an interesting conversation still had power to charm him but often during its continuance the full consciousness of his misfortune forced itself upon his mind for the majority of the subjects discussed by the artists came to them through the medium of sight and referred to new creations of architecture sculpture and painting from whose enjoyment his blindness \textcolor{green!50!black}{\textbf{debarred}} him

Gold: true an interesting conversation still had power to charm him but often during its continuance the full consciousness of his misfortune forced itself upon his mind for the majority of the subjects discussed by the artists came to them through the medium of sight and referred to new creations of architecture sculpture and painting from whose enjoyment his blindness \textbf{debarred} him

\medskip
\hrule
\medskip

\begin{center}
\colorbox{red!10}{\strut\textbf{\textcolor{red!70!black}{Incorrect Hybrid edits}}}
\end{center}

Beam Search: he forgot as sergey ivanovitch explained to him afterwards this syllogism that it was necessary for the public good to get rid of the marshal of the province that to get rid of the marshal it was necessary to have a majority of votes that to get a majority of votes it was necessary to secure \textcolor{green!50!black}{\textbf{flerov}} is right to vote that to secure the recognition of \textcolor{green!50!black}{\textbf{flerov}} is right to vote they must decide on the interpretation to be put on the act

Hybrid Search $\tau=\frac{1}{2}$: he forgot as sergey ivanovitch explained to him afterwards this syllogism that it was necessary for the public good to get rid of the marshal of the province that to get rid of the marshal it was necessary to have a majority of votes that to get a majority of votes it was necessary to secure \textcolor{red}{\textbf{fleurov}} is right to vote that to secure the recognition of \textcolor{red}{\textbf{fleurov}} is right to vote they must decide on the interpretation to be put on the act

Gold: he forgot as sergey ivanovitch explained to him afterwards this syllogism that it was necessary for the public good to get rid of the marshal of the province that to get rid of the marshal it was necessary to have a majority of votes that to get a majority of votes it was necessary to secure \textbf{flerov} is right to vote that to secure the recognition of \textbf{flerov} is right to vote they must decide on the interpretation to be put on the act

\medskip

Beam Search: why before he looked like the orneryest old rip that ever was but now when he would take off his new white beaver and make a bow and do a smile he looked that grand and good and pious that you would say he had walked right out of the ark and maybe was old \textcolor{green!50!black}{\textbf{leviticus}} himself

Hybrid Search $\tau=\frac{1}{2}$: why before he looked like the orneryest old rip that ever was but now when he would take off his new white beaver and make a bow and do a smile he looked that grand and good and pious that you would say he had walked right out of the ark and maybe was old \textcolor{red}{\textbf{leviathus}} himself

Gold: why before he looked like the orneriest old rip that ever was but now when he would take off his new white beaver and make a bow and do a smile he looked that grand and good and pious that you would say he had walked right out of the ark and maybe was old \textbf{leviticus} himself

\medskip

Beam Search: she was perfectly safe after writing to basil \textcolor{green!50!black}{\textbf{ransom}} and indeed it was difficult to see what he could have done to her except thank her he was only exceptionally superlative for her letter and assure her that he would come and see her the 1st time his business he was beginning to get a little should take him to boston

Hybrid Search $\tau=\frac{1}{2}$: she was perfectly safe after writing to basil \textcolor{red}{\textbf{ransome}} and indeed it was difficult to see what he could have done to her except thank her he was only exceptionally superlative for her letter and assure her that he would come and see her the 1st time his business he was beginning to get a little should take him to boston

Gold: she was perfectly safe after writing to basil \textbf{ransom} and indeed it was difficult to see what he could have done to her except thank her he was only exceptionally superlative for her letter and assure her that he would come and see her the 1st time his business he was beginning to get a little should take him to boston
\end{tcolorbox}

\begin{tcolorbox}[width=\columnwidth, enhanced, breakable]
\small
\setlength{\parindent}{0pt}

\textbf{[SPGISpeech]}

\begin{center}
\colorbox{green!12}{\strut\textbf{\textcolor{green!50!black}{Correct Hybrid edits}}}
\end{center}

Beam Search: i think what we have found over a \textcolor{red}{\textbf{multi year}} period here is that we often come across great ideas that neither provide the liquidity nor the time horizon that allow for us to be able to make it much of an investment in some of the larger pooled assets that we run

Hybrid Search $\tau=\frac{1}{2}$: i think what we have found over a \textcolor{green!50!black}{\textbf{multiyear}} period here is that we often come across great ideas that neither provide the liquidity nor the time horizon that allow for us to be able to make it much of an investment in some of the larger pooled assets that we run

Gold: i think what we have found over a \textbf{multiyear} period here is that we often come across great ideas that neither provide the liquidity nor the time horizon that allow for us to be able to make it much of an investment in some of the larger pooled assets that we run

\medskip

Beam Search: \textcolor{red}{\textbf{actuarial}} results may differ materially from the results expressed or implied in these statements as a result of risks uncertainties and other factors included but not limited to the factors set forth in the company is filings with the securities and exchange commission

Hybrid Search $\tau=\frac{1}{2}$: \textcolor{green!50!black}{\textbf{actual}} results may differ materially from the results expressed or implied in these statements as a result of risks uncertainties and other factors included but not limited to the factors set forth in the company is filings with the securities and exchange commission

Gold: \textbf{actual} results may differ materially from the results expressed or implied in these statements as a result of risks uncertainties and other factors including but not limited to the factors set forth in the company is filings with the securities and exchange commission

\medskip

Beam Search: if you take a week of our \textcolor{red}{\textbf{us}} business it is 15000000 of orders now even if you say that we were stopped for the full week and we lost 15000000 of orders which has not happened we were not stopped for the full week we did not lose 15000000 of orders

Hybrid Search $\tau=\frac{1}{2}$: if you take a week of our \textcolor{green!50!black}{\textbf{u s}} business it is 15000000 of orders now even if you say that we were stopped for the full week and we lost 15000000 of orders which has not happened we were not stopped for the full week we did not lose 15000000 of orders

Gold: if you take a week of our \textbf{u s} business it is 15000000 of orders now even if you say that we were stopped for the full week and we lost 15000000 of orders which has not happened we were not stopped for the full week we did not lose 15000000 of orders

\medskip
\hrule
\medskip

\begin{center}
\colorbox{red!10}{\strut\textbf{\textcolor{red!70!black}{Incorrect Hybrid edits}}}
\end{center}

Beam Search: for a protocol violation after he received an unrelated vaccination for shingles just before his 2nd dose of \textcolor{green!50!black}{\textbf{sel}} 212 his uric acid level had been fully controlled over the 1st month

Hybrid Search $\tau=\frac{1}{2}$: for a protocol violation after he received an unrelated vaccination for shingles just before his 2nd dose of \textcolor{red}{\textbf{sele}} 212 his uric acid level had been fully controlled over the 1st month

Gold: for a protocol violation after he received an unrelated vaccination for shingles just before his 2nd dose of \textbf{sel} 212 his uric acid level had been fully controlled over the 1st month

\medskip

Beam Search: the programs have so far generated more than 1300 ideas and initiatives with individual initiatives benefits ranging from 1000s of dollars to multimillions over fiscal year \textcolor{green!50!black}{\textbf{17}} and beyond

Hybrid Search $\tau=\frac{1}{2}$: the programs have so far generated more than 1300 ideas and initiatives with individual initiatives benefits ranging from 1000s of dollars to multimillions over fiscal year \textcolor{red}{\textbf{2017}} and beyond

Gold: the programs has so far generated more than 1300 ideas and initiatives with individual initiatives benefits ranging from 1000s of dollars to multimillions over fiscal year \textbf{17} and beyond

\medskip

Beam Search: from an ongoing ebit standpoint it is not really going to have an impact and as far as further reducing costs in that business all those costs were really contained in \textcolor{green!50!black}{\textbf{irapuato}} and there was very little cost here to deal with that particular issue so i do not think we are going to see anything

Hybrid Search $\tau=\frac{1}{2}$: from an ongoing ebit standpoint it is not really going to have an impact and as far as further reducing costs in that business all those costs were really contained in \textcolor{red}{\textbf{rio tinto}} and there was very little cost here to deal with that particular issue so i do not think we are going to see anything

Gold: from an ongoing ebit standpoint it is not really going to have an impact as far as further reducing costs on that business all those costs are really contained in \textbf{irapuato} and there was very little cost here dealing with that particular issue so i do not think we are going to see anything
\end{tcolorbox}

\begin{tcolorbox}[width=\columnwidth, enhanced, breakable]
\small
\setlength{\parindent}{0pt}

\textbf{[TED-LIUM]}

\begin{center}
\colorbox{green!12}{\strut\textbf{\textcolor{green!50!black}{Correct Hybrid edits}}}
\end{center}

Beam Search: making it can you see it yes good this is actually me making the i am not good at life face this is a piece of graffiti in my old neighborhood in berkeley california where i did my \textcolor{red}{\textbf{ph d}} on why we are better in games than we are in real life and this is a problem that a lot of gamers have we feel that we are not as good in reality as we are in games and i do not mean just good as in successful although that is part of it we do achieve more in game worlds but i also mean good as in motivated to do something that matters inspired to collaborate and to cooperate when we are in game worlds

Hybrid Search $\tau=\frac{1}{2}$: making it can you see it yes good this is actually me making the i am not good at life face this is a piece of graffiti in my old neighborhood in berkeley california where i did my \textcolor{green!50!black}{\textbf{phd}} on why we are better in games than we are in real life and this is a problem that a lot of gamers have we feel that we are not as good in reality as we are in games and i do not mean just good as in successful although that is part of it we do achieve more in game worlds but i also mean good as in motivated to do something that matters inspired to collaborate and to cooperate when we are in game worlds

Gold: making it can you see yes good this is actually me making the i am not good at life face this is a piece of graffiti in my old neighborhood in berkeley california where i did my \textbf{phd} on why we are better in games than we are in real life and this is a problem that a lot of gamers have we feel that we are not as good in reality as we are in games and i do not mean just good as in successful although that is part of it we do achieve more in game worlds but i also mean good as in motivated to do something that matters inspired to collaborate and to cooperate and when we are in game worlds

\medskip

Beam Search: and the fundamental lesson i believe is that design truly is a contact sport it demands that we bring all of our senses to the task and that we apply the very best of our thinking our feeling and our doing to the challenge that we have at hand and sometimes a little prototype of this experience is all that it takes to turn us from an ooh 0 moment to a \textcolor{red}{\textbf{tada}} moment and that can make a big difference thank you very much

Hybrid Search $\tau=\frac{1}{2}$: and the fundamental lesson i believe is that design truly is a contact sport it demands that we bring all of our senses to the task and that we apply the very best of our thinking our feeling and our doing to the challenge that we have at hand and sometimes a little prototype of this experience is all that it takes to turn us from an ooh 0 moment to a \textcolor{green!50!black}{\textbf{ta da}} moment and that can make a big difference thank you very much

Gold: and the fundamental lesson i believe is that design truly is a contact sport it demands that we bring all of our senses to the task and that we apply the very best of our thinking our feeling and our doing to the challenge that we have at hand and sometimes a little prototype of this experience is all that it takes to turn us from an 0 moment to a \textbf{ta da} moment and that can make a big difference thank you very much

\medskip

Beam Search: bill gross has several companies including one called e solar that has some great solar thermal technology vinod \textcolor{red}{\textbf{kolstad}} is investing in dozens of companies that are doing great things and have interesting possibilities and i am i am trying to help back that and nathan meriwald and i actually are backing a company

Hybrid Search $\tau=\frac{1}{2}$: bill gross has several companies including one called e solar that has some great solar thermal technology vinod \textcolor{green!50!black}{\textbf{khosla}} is investing in dozens of companies that are doing great things and have interesting possibilities and i am i am trying to help back that and nathan meriwald and i actually are backing a company

Gold: bill gross has several companies including one called esolar that has some great solar thermal technologies vinod \textbf{khosla} is investing in dozens of companies that are doing great things and have interesting possibilities and i am i am trying to help back that nathan myhrvold and i actually are backing a company

\medskip
\hrule
\medskip

\begin{center}
\colorbox{red!10}{\strut\textbf{\textcolor{red!70!black}{Incorrect Hybrid edits}}}
\end{center}

Beam Search: this guy was a hero jonas saw he took one of the worst scourges of mankind away from us no fear no agony polio \textcolor{green!50!black}{\textbf{puff}} gone

Hybrid Search $\tau=\frac{1}{2}$: this guy was a hero jonas saw he took one of the worst scourges of mankind away from us no fear no agony polio \textcolor{red}{\textbf{poof}} gone

Gold: this guy was a hero jonas salk he took one of the worst scourges of mankind away from us no fear no agony polio \textbf{puff} gone

\medskip

Beam Search: \textcolor{green!50!black}{\textbf{and}} i am telling a story that many of you know because steve is columns became the basis for a book which was turned into a movie

Hybrid Search $\tau=\frac{1}{2}$: i am telling a story that many of you know because steve is columns became the basis for a book which was turned into a movie

Gold: \textbf{and} i am telling a story that many of you know because steve is columns became the basis for a book which was turned into a movie

\medskip

Beam Search: i would like to leave you with a poem by a 14th century persian poet named \textcolor{green!50!black}{\textbf{hafiz}}

Hybrid Search $\tau=\frac{1}{2}$: i would like to leave you with a poem by a 14th century persian poet named \textcolor{red}{\textbf{hafez}}

Gold: i would like to leave you with a poem by a 14th century persian poet named \textbf{hafiz}
\end{tcolorbox}

\begin{tcolorbox}[width=\columnwidth, enhanced, breakable]
\small
\setlength{\parindent}{0pt}

\textbf{[VoxPopuli]}

\begin{center}
\colorbox{green!12}{\strut\textbf{\textcolor{green!50!black}{Correct Hybrid edits}}}
\end{center}

Beam Search: \textcolor{red}{\textbf{cedaw}} is role is to contribute to the shaping and implementing of vet but also shaping skills and qualification policies at the union level the agency can do this by providing crucial evidence and services for policy making and knowledge sharing amongst the union and national actors in particular to governments and social partners i believe that the revision of the functioning of this tripartite agency has been a very good one and i fully appreciate the agency is great work in supporting and developing of inclusive and quality vet systems now we have to work together that we put in place

Hybrid Search $\tau=\frac{1}{2}$: \textcolor{green!50!black}{\textbf{cedefop}} is role is to contribute to the shaping and implementing of vet but also shaping skills and qualification policies at the union level the agency can do this by providing crucial evidence and services for policy making and knowledge sharing amongst the union and national actors in particular to governments and social partners i believe that the revision of the functioning of this tripartite agency has been a very good one and i fully appreciate the agency is great work in supporting and developing of inclusive and quality vet systems now we have to work together that we put in place

Gold: \textbf{cedefop} is role is to contribute to the shaping and implementing of vt but also to shaping skills and qualification policies at union level the agency can do this by providing crucial evidence and services for policymaking and knowledge sharing among the union and national actors in particular to governments and social partners i believe that the revision of the function of this tripartite agency has been a very good one and i fully appreciate the agency is great work in supporting and developing inclusive and quality vt systems now we have to work together to put in place

\medskip

Beam Search: eu is intention is to help the yemeni population and this is why we need to support the efforts in identifying a political solution as soon as possible using dialog and negotiations we need to understand that all the parties in the conflict must comply with their obligations under international law

Hybrid Search $\tau=\frac{1}{2}$: \textcolor{green!50!black}{\textbf{the}} eu is intention is to help the yemeni population and this is why we need to support the efforts in identifying a political solution as soon as possible using dialog and negotiations we need to understand that all the parties in the conflict must comply with their obligations under international law

Gold: \textbf{the} eu is intention is to help the yemeni population and this is why we need to support the efforts in identifying a political solution as soon as possible using dialog and negotiations we need to understand that all parties in the conflict must comply with their obligations under international law

\medskip

Beam Search: eu aid worth 1000000s of euros was destroyed in full impunity and with no adequate labeling of settlement products the village of \textcolor{red}{\textbf{kanalmach}} that i and countless friends of the bedouins have visited is about to be destroyed

Hybrid Search $\tau=\frac{1}{2}$: eu aid worth 1000000s of euros was destroyed in full impunity and with no adequate labeling of settlement products the village of \textcolor{green!50!black}{\textbf{khan al ahmar}} that i and countless friends of the bedouins have visited is about to be destroyed

Gold: eu aid worth 1000000s of euros destroyed in full impunity and no adequate labeling of settlement products the village of \textbf{khan al ahmar} that i and countless friends of the bedouins have visited is about to be destroyed

\medskip
\hrule
\medskip

\begin{center}
\colorbox{red!10}{\strut\textbf{\textcolor{red!70!black}{Incorrect Hybrid edits}}}
\end{center}

Beam Search: but we must also set clear conditions most important of all is the return of the rule of law and political freedoms and the \textcolor{green!50!black}{\textbf{re establishment}} of civilian rule over the military it must also include a clear path to inclusive and fair elections in 2018 which could also be observed by the european parliament

Hybrid Search $\tau=\frac{1}{2}$: but we must also set clear conditions most important of all is the return of the rule of law and political freedoms and the \textcolor{red}{\textbf{reestablishment}} of civilian rule over the military it must also include a clear path to inclusive and fair elections in 2018 which could also be observed by the european parliament

Gold: but we must also set clear conditions most important of all is the return of the rule of law and political freedoms and the \textbf{re establishment} of civilian rule over the military it must also include a clear path to inclusive and fair elections in 2018 which could also be observed by the european parliament moreover

\medskip

Beam Search: i would also like to say that the position of the president of the commission in some ways is understandable \textcolor{green!50!black}{\textbf{but}} on the other hand there is a huge risk in giving up the community method and coming back to national contributions in special instruments

Hybrid Search $\tau=\frac{1}{2}$: i would also like to say that the position of the president of the commission in some ways is understandable \textcolor{red}{\textbf{understandable}} on the other hand there is a huge risk in giving up the community method and coming back to national contributions in special instruments

Gold: i would like to say as well that the position of the president of the commission is in some ways understandable \textbf{but} on the other hand there is a huge risk concerning giving up the community method and going back to national contributions in special instruments

\medskip

Beam Search: memory herself has said that marriage is often the end for girls like me but if our leaders will invest in us and give us the chance to be educated we will become women who create a better society for everyone

Hybrid Search $\tau=\frac{1}{2}$: \textcolor{red}{\textbf{madam president}} memory herself has said that marriage is often the end for girls like me but if our leaders will invest in us and give us the chance to be educated we will become women who create a better society for everyone

Gold: memory herself has said marriage is often the end for girls like me but if our leaders will invest in us and give us the chance to be educated we will become women who create a better society for everyone
\end{tcolorbox}

\subsection{Layer-wise Features Visualization}
\label{apx:heat_map}

\begin{figure*}[t]
\centering
\includegraphics[width=\textwidth]{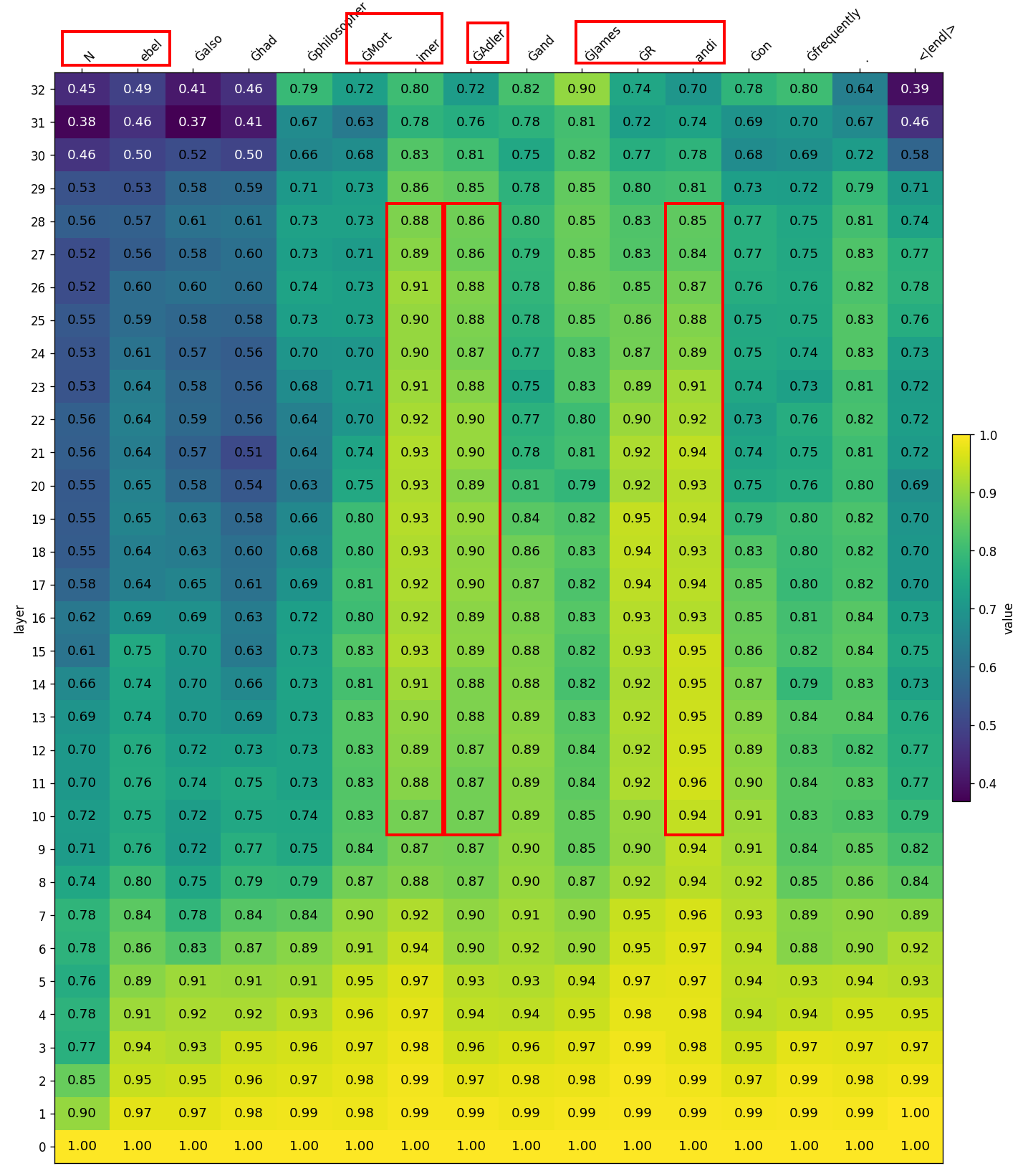}
\caption{
Layer-wise heatmaps for \texttt{Common Voice} showing AngleFeat values. The text reads: ``Nebel also had philosopher Mortimer Adler and James Randi on frequently.'' Named entities ``Nebel'', ``Mortimer'', ``Adler'', ``James'', and ``Randi'' are highlighted by the red box. Some of the named-entity tokens exhibit higher $\textrm{AngleFeat}$ values.
}
\label{fig:heatmap-common-voice-cos}
\end{figure*}

\begin{figure*}[t]

\includegraphics[width=\textwidth]{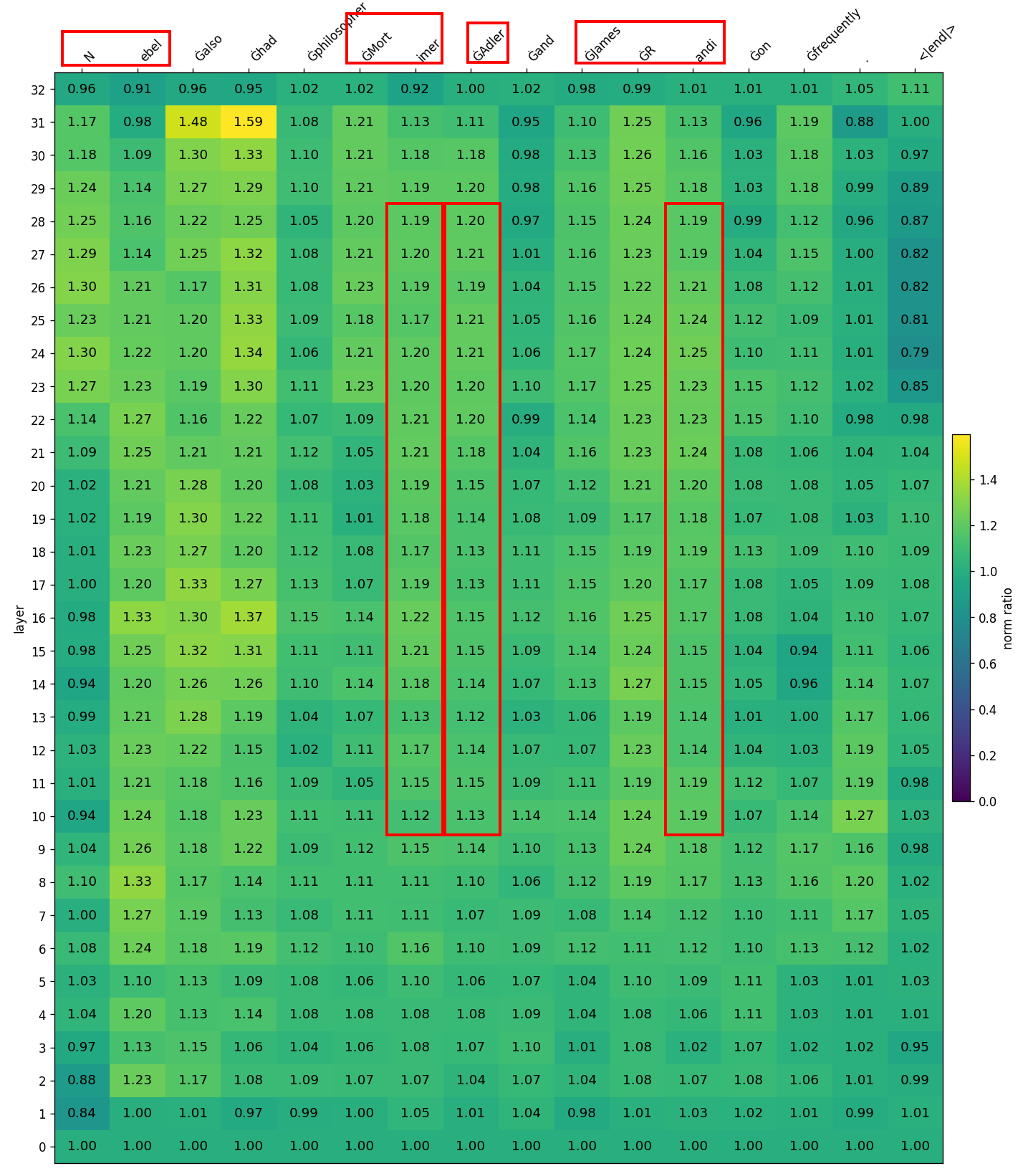}

\caption{
Layer-wise heatmaps for \texttt{Common Voice} showing $\textrm{MagFeat}$ values. The text reads: ``Nebel also had philosopher Mortimer Adler and James Randi on frequently.'' Named entities ``Nebel'', ``Mortimer'', ``Adler'', ``James'', and ``Randi'' are highlighted by the red box. Some of the named-entity tokens exhibit higher $\textrm{MagFeat}$ values.
}
\label{fig:heatmap-common-voice-norm}
\end{figure*}

We show heatmaps of the layer-wise features for a Common Voice example in Figure~\ref{fig:heatmap-common-voice-cos} and Figure~\ref{fig:heatmap-common-voice-norm}. Named-entity words such as ``Mortimer'', ``Adler'' and ``Randi'' exhibit higher cosine similarities and norm ratios in intermediate layers.


\end{document}